\documentclass[letterpaper]{article} 
\usepackage{aaai24}  
\usepackage{times}  
\usepackage{helvet}  
\usepackage{courier}  
\usepackage[hyphens]{url}  
\usepackage{graphicx} 
\usepackage{natbib}  
\usepackage{caption} 
\usepackage{algorithm}
\usepackage{algorithmic}

\usepackage{newfloat}
\usepackage{listings}

\usepackage{soul}
\usepackage{url}
\usepackage[utf8]{inputenc}
\usepackage{amsmath}
\usepackage{adjustbox}
\usepackage{caption}
\usepackage{algorithm}
\usepackage{algorithmic}
\usepackage{color}
\usepackage[switch]{lineno}
\usepackage{multirow}
\usepackage{makecell}
\usepackage{pifont}
\usepackage{amssymb}

\usepackage{colortbl}
\usepackage{xcolor}
\usepackage{booktabs}
\usepackage{array}
\usepackage{arydshln} 

\definecolor{lightgray}{gray}{0.85}
\definecolor{darkgray}{gray}{0.5}
\sethlcolor{lightgray}

\newcommand{\red}[0]{\textcolor[rgb]{ .753,  0,  0}}
\newcommand{\tif}[1]{\textit{\textbf{#1}}}

\usepackage{paralist} 
\usepackage{bm}

\newcommand{\proposed}[0]{SGID}

\newcommand{\eg}[0]{\textit{e.g.},}
\newcommand{\ie}[0]{\textit{i.e.},}

\DeclareCaptionStyle{ruled}{labelfont=normalfont,labelsep=colon,strut=off} 
\floatstyle{ruled}
\newfloat{listing}{tb}{lst}{}
\floatname{listing}{Listing}
\title{Semantic-Guided Generative Image Augmentation Method \\ with Diffusion Models for Image Classification}
\author{
    Bohan Li, 
    Xiao Xu,
    Xinghao Wang,
    Yutai Hou,
    Yunlong Feng,\\
    Feng Wang,
    Xuanliang Zhang,
    Qingfu Zhu\thanks{Corresponding Authors.},
    Wanxiang Che
}
\affiliations{
    Harbin Institute of Technology, Harbin, China\\
    \{bhli, xxu, xhwang, ythou, ylfeng\}@ir.hit.edu.cn, \{7203610216, 120l020412\}@stu.hit.edu.cn, \{qfzhu, car\}@ir.hit.edu.cn
}

\usepackage{bibentry}

\begin{document}

\maketitle

\begin{abstract}

  Existing image augmentation methods consist of two categories: perturbation-based methods and generative methods.
  Perturbation-based methods apply pre-defined perturbations to augment an original image, but only locally vary the image, thus lacking image diversity.
  In contrast, generative methods bring more image diversity in the augmented images but may not preserve semantic consistency, thus may incorrectly changing the essential semantics of the original image. 
  To balance image diversity and semantic consistency in augmented images, we propose \textbf{\proposed{}}, a \textbf{S}emantic-guided \textbf{G}enerative \textbf{I}mage augmentation method with \textbf{D}iffusion models for image classification. 
  Specifically, \proposed{} employs diffusion models to generate augmented images with good image diversity. 
  More importantly, \proposed{} takes image labels and captions as guidance to maintain semantic consistency between the augmented and original images.
  Experimental results show that \proposed{} outperforms the best augmentation baseline by $1.72\%$ on ResNet-50 (from scratch), $0.33\%$ on ViT (ImageNet-21k), and $0.14\%$ on CLIP-ViT (LAION-2B). 
  Moreover, \proposed{} can be combined with other image augmentation baselines and further improves the overall performance.
  We demonstrate the semantic consistency and image diversity of \proposed{} through quantitative human and automated evaluations, as well as qualitative case studies. 
  
\end{abstract}

\section{Introduction}\label{sec:intro}

The data-hungry problem in deep learning
has been a hot topic~\cite{minaee2021image} since a sufficient number of training samples is crucial for unleashing the power of deep networks~\cite{zhang2022expanding}. 
However, manually collecting and labeling large-scale datasets are both expensive and time-consuming~\cite{LI_DA_surevey}, giving rise to the Data Augmentation (DA) methods.
Take image classification as an example, generally, a DA method generates augmented images that are diverse from the original training images while preserving the essential semantics of the original images.
It has been demonstrated to be effective in improving the model performance on classification tasks in practice~\cite{Dunlap2023DiversifyYV}.

\begin{figure}[!t]
  \centering 
  \includegraphics[width=1.00 \linewidth]{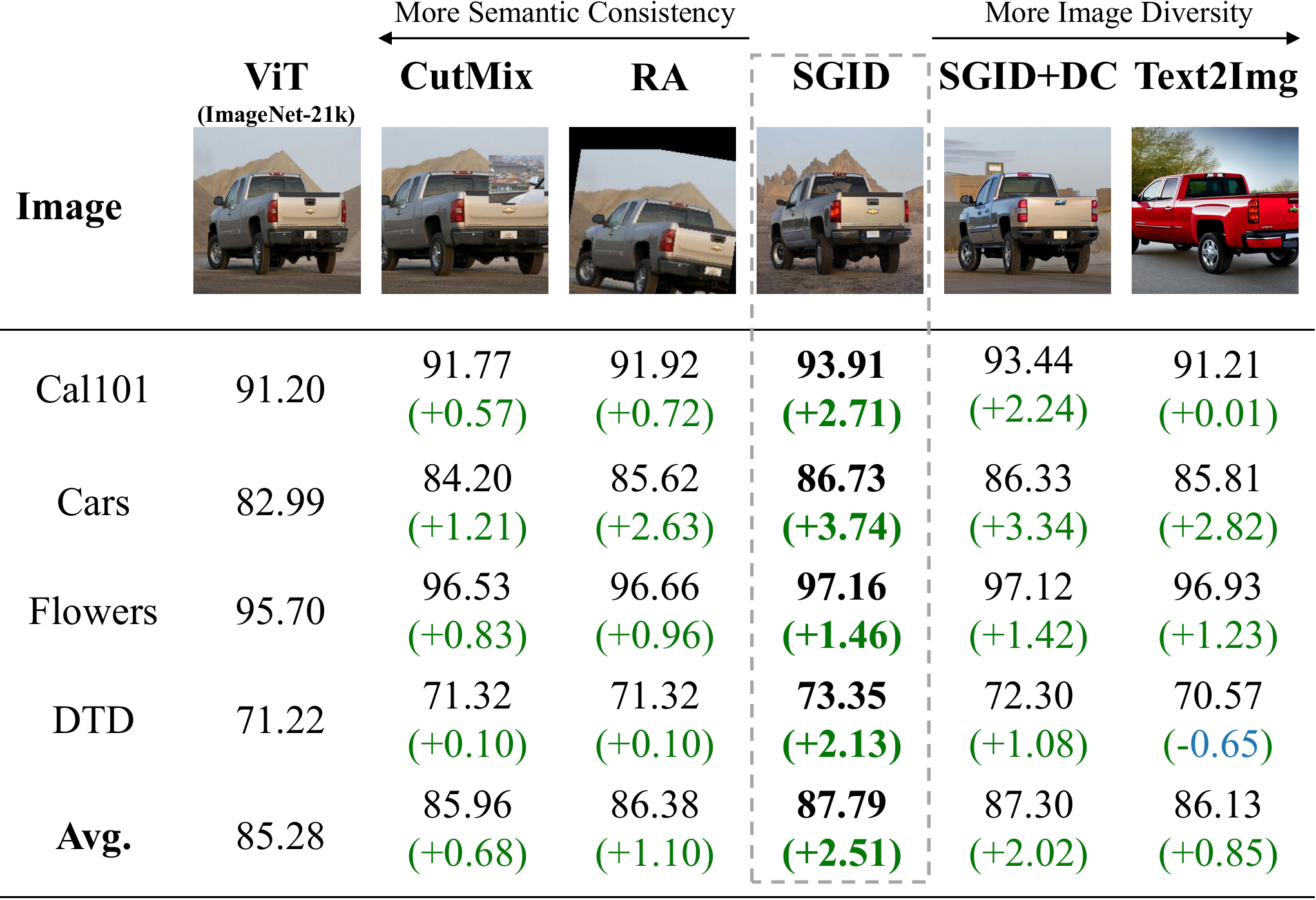}
  \caption{
  A comparison of four baseline methods and our proposed \proposed{} using the ViT (ImageNet-21k) backbone across four datasets. Our \proposed{} strikes a balance between semantic consistency and image diversity, leading to the highest performance improvement.
  }
  \label{fig:intro-compare}
\end{figure}

Typically, DA methods for image classification can be divided into two categories: perturbation-based and generation-based.
The former obtains augmented images by modifying the original image with pre-defined perturbations, \eg{} image erasing~\cite{zhong2020random} and image  mixup~\cite{yun2019cutmix}.
In this way, most of the semantics remain since the augmented image only locally differs from the original image in a limited area.
However, the diversity is quite limited at the same time and becomes the bottleneck of the methods of this category.
In contrast, the generation-based methods synthesize augmented images by generative models like diffusion models~\cite{rombach2022high}.
They directly generate augmented images based on label-related captions and/or original images.
In this way, they can generate quite diverse images but is inferior to the perturbation-based methods in preserving semantics, which are less diverse but more semantically consistent.
The excessive noise or diversity introduced by generative DA methods may \textbf{incorrectly change} the essential semantics of the original image. 
As shown in Figure~\ref{fig:intro-compare}, the last two augmented images incorrectly show two door handles on the same side of the door or change the badge on the rear of the car.

Naturally, we intend to explore how to maintain a balance between image diversity and semantic consistency in a generative image DA method, which has not yet been systematically explored by existing methods, to achieve better performance on image classification tasks.
To this end, we propose \proposed{}, a \textbf{S}emantic-guided \textbf{G}enerative \textbf{I}mage augmentation method with \textbf{D}iffusion models for image classification.
It ensures semantic consistency between the original and augmented images, and achieves good image diversity (see Figure~\ref{fig:intro-compare}).
\proposed{} consists of two steps:
\tif{(1)} Collect the text label of the image and then generate the image caption with the BLIP~\cite{li2022blip} model. Both of them convey the essential semantics of the original image.
\tif{(2)} Concatenate the label and caption to construct a textual prompt and subsequently feed it into the Stable Diffusion~\cite{rombach2022high} model along with the original image. 
\proposed{} tends to generate diverse augmented images, while the textual prompt as semantic guidance help to preserve the essential semantics of the original images (see Sec~\ref{sec:diverdity}).

We conduct experiments on seven image classification datasets based on three backbones including ResNet-50 (from scratch)~\cite{he2016deep}, ViT (ImageNet-21k)~\cite{dosovitskiy2020image}, and CLIP-ViT (LAION-2B)~\cite{Cherti2022ReproducibleSL}.
We compare \proposed{} with seven strong DA baselines including four perturbation-based methods and three generative methods.
Our method outperforms the backbones on all datasets and achieves the best or comparable performance to all baselines. 
Especially, \proposed{} leads to $10.39\%, 2.08\%, 0.85\%$ average accuracy gain across three backbones and seven datasets, respectively.
Moreover, we find that combining \proposed{} with standard DA baselines can achieve further improvement on seven datasets.
We further compare the semantic consistency and diversity between \proposed{} as well as baselines by human evaluation, automatic similarity evaluation, and case studies.

Our contributions are as follows:
\begin{compactitem}
\item We propose a \textbf{S}emantic-guided \textbf{G}enerative \textbf{I}mage augmentation method with \textbf{D}iffusion models (\proposed{}) for image classification. 
Human evaluation, automated evaluations, and case studies demonstrate that \proposed{} balances the semantic consistency and image diversity.
\item \proposed{} achieves better average performance than the best baseline by $1.72\%$, $0.33\%$, and $0.14\%$ across three backbones and seven datasets.
\item \proposed{} can be combined with other image augmentation baselines and further improves the overall performance. 
\end{compactitem}

\section{Background}\label{sec:background}
Image augmentation is widely used in computer vision~\cite{minaee2021image,algan2021image}. 
It generates augmented images that are diverse from the original images while preserving their semantics.
Common augmentation methods include \tif{(1)} perturbation-based methods and \tif{(2)} generative methods.

Perturbation-based methods apply pre-defined perturbations to augment an original image and preserve image semantic consistency~\cite{yang2022image}.
\textit{Random Erasing}~\cite{zhong2020random}
generates augmented images by deleting one or more subregions of an image.
\textit{CutMix}~\cite{yun2019cutmix}
uses the full or partial features and labels of different images to apply some interpolation methods.
\textit{RandAugment}~\cite{cubuk2020randaugment}
tries to search the space of augmentation methods according to different tasks.
\textit{MoEx}~\cite{li2021feature}
performs the transformation in a learned feature space rather than conducting augmentation only in the input space.
Despite their effectiveness in image classification, the pre-defined perturbations of these methods only locally vary the images, thus lacking diversity.

Thanks to the development of generative models like Stable Diffusion (SD), an image generation model pre-trained on large-scale image-text pairs, there are some attempts at generative methods for more diversity~\cite{zhang2022expanding}.
\textit{Text2Img}~\citep{He2022IsSD} applies a fine-tuned T5 model
to generate captions based on image labels, and then employs a text-to-image diffusion model to generate images without using the original image.
\citet{Dunlap2023DiversifyYV} respectively use diverse captions or editing instructions to modify original images into augmented images. 
During image generation, they correspondingly employ the Image2Image diffusion model or the InstructPix2Pix diffusion model~\cite{brooks2023instructpix2pix} as generative models.
Inspired by this work, we take its spirit into our \proposed{} to obtain two variants of \proposed{} as generative baselines named
\textit{\proposed{}+DiverseCaption} and \textit{\proposed{}+InstructPix2Pix}.\footnote{We further discuss this work \cite{Dunlap2023DiversifyYV} and other generative methods in Appendix Sec.~\ref{sec:appendix-generative-method} and \ref{sec:appendix-gif}.}

However, existing generative methods
may incorrectly change the essential semantics of the original image, which is crucial for image classification~\cite{Burg2023ADA}.

This paper aims to improve the semantic consistency of generative methods through guiding the generation of augmented images by explicitly using the essential semantics of original images.
\proposed{} balances the image diversity and semantic consistency in augmented images, and achieves consistent performance gains across datasets and backbones.

\begin{figure*}[!t]
  \centering 
  \includegraphics[width=0.95\linewidth]{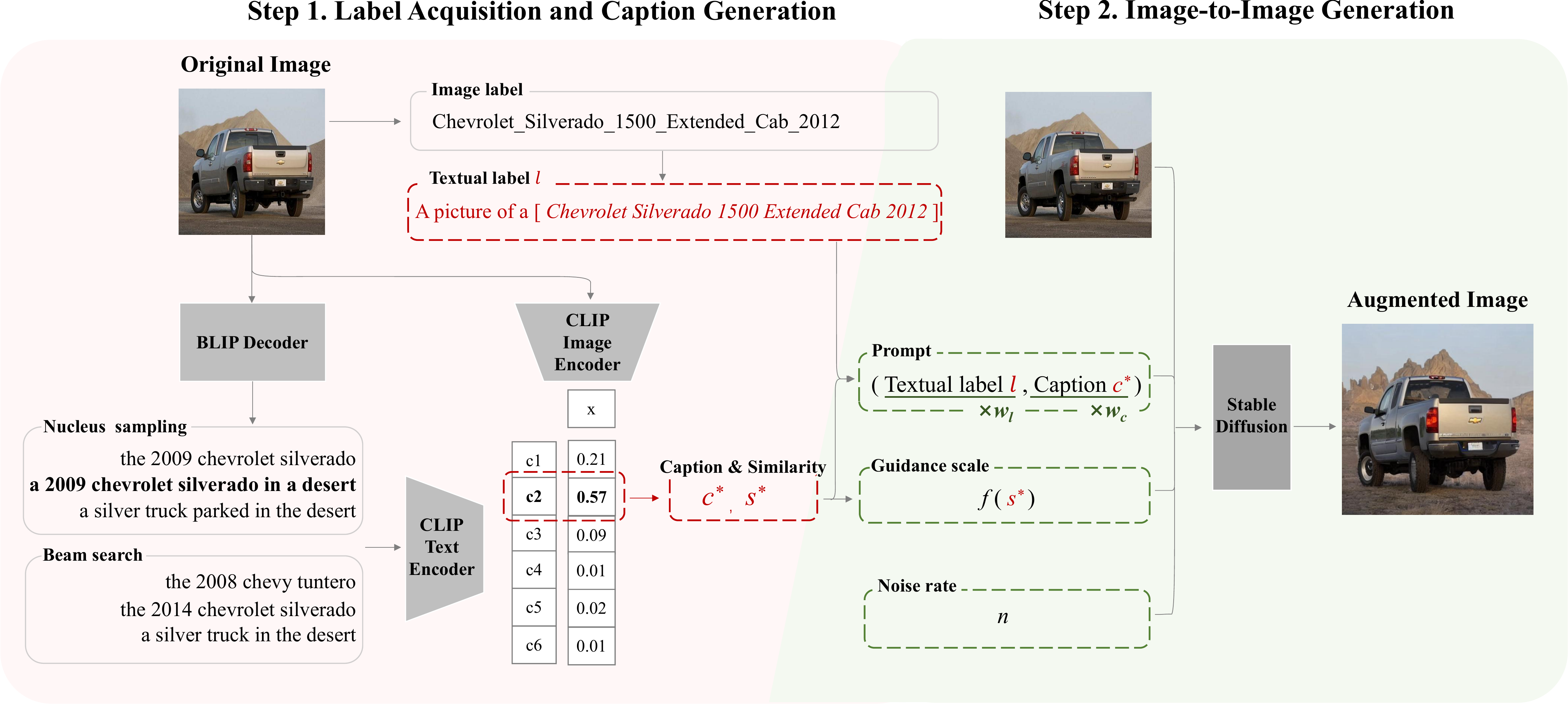}
  \caption{An illustration of our \textbf{\proposed{}}. \textbf{Step 1} first collects textual labels for each image, then use BLIP to generate captions, and then use CLIP to calculate the similarity between the chosen caption and the original image. \textbf{Step 2} generates the augmented images through the Stable Diffusion model, utilizing the original image, the prompt consisting of the textual label and caption, the noise rate, and the guidance scale based on the similarity.}
  \label{fig:main_method}
\end{figure*}

\section{\proposed{}}\label{Sec.method}
In this paper, we propose \proposed{}, 
a semantic-guided generative image augmentation method with diffusion models for image classification.
We do not aim to exceed existing image augmentation baselines on various datasets and different backbones, but to explore such a method that balances image diversity and semantic consistency, \ie{}
preserving the essential semantics of the original images and simultaneously bringing good image diversity.
In addition, \proposed{} can be naturally combined with other DA baselines and further improve their performance.
It consists of two essential steps, as illustrated in Figure~\ref{fig:main_method}:
\begin{compactenum}
    \item We first collect textual labels for each image, then use BLIP to generate captions, and then use CLIP~\cite{radford2021learning} to calculate the similarity between the chosen caption and the original image. Both labels and captions provide the essential semantics of the original images.
    \item We construct the prompt based on the label and the caption of each image.
    The prompt is subsequently fed into Stable Diffusion along with the original image.
    The semantic guidance contained in the prompt can help generate diverse and semantic-consistent augmented images.\footnote{The introduction to BLIP, CLIP, and SD is Appendix Sec.~\ref{sec:appendix-PMS}.} 
\end{compactenum}

\subsection{Label Acquisition and Caption Generation}

In \textbf{Step 1}, we construct prompts from original images as semantic guidance for the subsequent \textbf{Step 2}. 
The prompt consists of the textual label and the caption of the original image. 
We argue that the image label contains accurate semantic information and the caption provides the overall description of the image.
Thus, the prompt is an explicit constraint to preserve semantic consistency.
Specifically, for each image ${x} \in \mathcal{X}$ in an image classification dataset $\mathcal{D} = (\mathcal{X}, \mathcal{Y})$, we first employ its groundtruth image label to construct a corresponding sentence as the textual label $l$:
\begin{equation}\label{textual-prompt}
  l = ``\texttt{A picture of a }\texttt{[label]}''.
\end{equation}
Then we generate captions for the image by the BLIP model:
\begin{equation}
\label{eq:gen_caption}
c = \texttt{BLIP}({x}).
\end{equation}

We explore different sampling strategies, including beam search~\cite{shen2017style} and nucleus sampling~\cite{Holtzman2019TheCC}).
Beam search tends to generate common and safe captions, hence offering less extra knowledge, and Nucleus sampling generates more diverse captions~\cite{li2022blip}.
We choose one of the sampling strategies based on the results of the validation set.
For each image, we obtain the corresponding caption set $\mathcal{C} = \{c_1, c_2, \ldots, c_k\}$ and randomly select one caption from it.
Captions can provide further semantic information beyond the object category of the image, such as the description of the background and color.
Combining captions with groundtruth but often too-short image labels can provide effective semantic guidance for image-to-image generation.

Moreover, to increase the semantic consistency of the prompt, we also explore potentially higher-quality captions by taking CLIP as an \textbf{optional} filter.
Given the generated caption set $\mathcal{C}$, we use CLIP to calculate the similarity between the original image with each caption and obtain the similarity set:  $\mathcal{S} = \{s_1, s_2, \ldots, s_k\}$. 
We select the caption with the highest similarity in $\mathcal{S}$.
We denote the selected caption and its similarity with ${c}^*$ and ${s}^*$. 
For example, in Figure~\ref{fig:main_method}, the caption ${c}^* = \text{``a 2009 chevrolet silverado in a desert''}$ generated by nucleus sampling gets the highest similarity ${s}^*=0.57$.

\begin{table*}[!ht]
  \centering
  \adjustbox{width=0.95\linewidth}{
    \begin{tabular}{clcccccccc}
    \toprule
          &       & \textbf{CIFAR-100} & \textbf{CIFAR-10} & \textbf{Cal101} & \textbf{Cars} & \textbf{Flowers} & \textbf{Pets} & \textbf{DTD} & \textbf{Avg.} \\
    \midrule
    \multirow{12}[8]{*}{ResNet-50 (from scratch)} & Backbone~\cite{he2016deep} & 75.12  & 94.81  & 47.38  & 82.26  & 33.02  & 50.30  & 19.85  & 57.53  \\
\cmidrule{2-10}          
          & + Random Erasing~\cite{zhong2020random} & 76.35  & 95.34  & 46.26  & 80.73  & 34.62  & 49.30  & 20.65  & 57.61  \\
          & + CutMix~\cite{yun2019cutmix} & \textbf{77.56} & \textbf{95.52} & 49.28  & 83.55  & 33.35  & 51.21  & 20.14  & 58.66  \\
          & + MoEx~\cite{li2021feature} & 76.00  & 95.23  & 52.74  & 81.13  & 36.27  & 55.79  & 21.20  & 59.77  \\
          & + RandAugment~\cite{cubuk2020randaugment} & \underline{76.40}  & 95.00  & \underline{58.55}  & 86.85  & 41.97  & 57.45  & 25.02  & 63.03  \\
\cdashline{2-10}[1pt/1pt]          
          & + Text2Img~\cite{He2022IsSD} & 74.99  & 94.97  & 57.60  & 85.44  & 42.06  & 67.08  & 31.18  & 64.76  \\
          & \red{+ \proposed{} (Ours)} & \red{75.72} & \red{\underline{95.48}} & \red{\textbf{59.17*}} & \red{\textbf{88.53*}} & \red{\textbf{45.61*}} & \red{\textbf{73.71*}} & \red{\textbf{37.19*}} & \red{\textbf{67.92*}} \\
          &  ~~~~  + DiverseCaption~\cite{Dunlap2023DiversifyYV} & 75.14  & 95.10  & 58.29  & \underline{87.86}  & \underline{43.66}  & \underline{69.73}  & \underline{33.61}  & \underline{66.20}  \\
          &  ~~~~  + InstructPix2Pix~\cite{Dunlap2023DiversifyYV} & 75.09  & 94.70  & 56.83  & 86.32  & 42.30  & 68.14  & 30.22  & 64.80  \\
\cmidrule{2-10}          
          & + \proposed{} \& MoEx & 77.54  & \underline{95.68}  & \underline{70.56}  & 87.33  & \underline{49.20}  & \underline{76.52}  & \underline{38.92}  & \underline{70.82}  \\
          & + \proposed{} \& RandAugment & \underline{77.56}  & 95.49  & \textbf{75.94} & \textbf{91.07} & \textbf{55.71} & \textbf{82.98} & \textbf{49.78} & \pmb{75.50} \\
          & + Text2Img \& \proposed{} & 75.40  & 95.06  & 64.20  & 87.94  & 42.36  & 69.28  & 35.79  & 67.15  \\
    \midrule
    \multirow{12}[8]{*}{ViT (ImageNet-21k)} & Backbone~\cite{dosovitskiy2020image} & 89.86  & 98.61  & 91.20  & 82.99  & 95.70  & 92.04  & 71.22  & 88.80  \\
\cmidrule{2-10}          
          & + Random Erasing~\cite{zhong2020random} & 89.87  & 98.67  &91.71  & 83.94  & 96.28  & 92.55  & 72.07  & 89.30  \\
          & + CutMix~\cite{yun2019cutmix} & 89.94  & 98.67  & 91.77  & 84.20  & 96.53  & 92.45  & 71.32  & 89.27  \\
          & + MoEx~\cite{li2021feature} & 90.11  & 98.67  & 92.24  & 85.94  & 96.78  & 92.66  & 70.79  & 89.60  \\
          & + RandAugment~\cite{cubuk2020randaugment} & 90.32  & 98.60  & 91.92  & 85.62  & 96.66  & 92.91  & 71.32  & 89.62  \\
\cdashline{2-10}[1pt/1pt]         
        & + Text2Img~\cite{He2022IsSD} & \textbf{92.66} & \textbf{99.01} & 91.21  & 85.81  & 96.93  & 92.67  & 70.57  & 89.84  \\
        & \red{+  \proposed{} (Ours)} & \red{\textbf{92.66}} & \red{\underline{98.96}} & \red{\textbf{93.91*}} & \red{\textbf{86.73*}} & \red{\textbf{97.16*}} & \red{\textbf{93.38}} & \red{\textbf{73.35*}} & \red{\textbf{90.88*}} \\
        & ~~~~+ DiverseCaption~\cite{Dunlap2023DiversifyYV} & \underline{92.59}  & 98.66  & \underline{93.44}  & \underline{86.33}  & \underline{97.12}  & \textbf{93.38 } & \underline{72.30}  & \underline{90.55}  \\
        & ~~~~+ InstructPix2Pix~\cite{Dunlap2023DiversifyYV} & 92.53  & 98.50  & 92.19  & 85.82  & 97.04  & \underline{93.02}  & 71.70  & 90.11  \\
\cmidrule{2-10}          
          & + \proposed{} \& MoEx & \underline{92.02}  & \underline{98.83}  & \underline{94.00}  & \underline{87.80}  & 96.69  & 92.99  & \underline{73.45}  & \underline{90.83}  \\
          & + \proposed{} \& RandAugment & 91.64  & 98.72  & \textbf{94.30} & \textbf{88.28} & \underline{97.13}  & \textbf{94.00} & \textbf{74.14} & \textbf{91.17} \\
          & + Text2Img \& \proposed{} & \textbf{92.69} & \textbf{99.03} & 91.35  & 85.84  & \textbf{97.17} & \underline{93.30}  & 71.27  & 90.09  \\
    \midrule
    \multirow{12}[8]{*}{CLIP-ViT (LAION-2B)} & Backbone~\cite{Cherti2022ReproducibleSL} & 85.89  & 95.04  & 92.96  & 85.13  & 91.46  & 93.69  & 65.81  & 87.14  \\
\cmidrule{2-10}          & + Random Erasing~\cite{zhong2020random} & \underline{86.26}  & 95.27  & \underline{94.32}  & 85.26  & 91.53  & 93.88  & 66.33  & 87.55  \\
          & + CutMix~\cite{yun2019cutmix} & 85.98  & 95.20  & 93.85  & 85.43  & 91.64  & \textbf{94.01} & 66.20  & 87.47  \\
          & + MoEx~\cite{li2021feature} & 86.02  & 95.16  & 93.81  & 85.79  & 91.97  & 93.76  & \textbf{67.18} & 87.67  \\
          & + RandAugment~\cite{cubuk2020randaugment} & 86.08  & 95.32  & 93.79  & 86.84  & 91.88  & 93.95  & \underline{66.74}  & 87.80  \\
\cdashline{2-10}[1pt/1pt]         & + Text2Img~\cite{He2022IsSD} & 85.80  & 94.77  & 93.72  & 86.80  & 91.90  & 93.71  & 64.50  & 87.31  \\
          & \red{+ \proposed{} (Ours)} & \red{\textbf{86.53*}} & \red{\textbf{95.66*}} & \red{94.29} & \red{\textbf{87.19*}} & \red{\textbf{92.04*}} & \red{\underline{93.98}} & \red{66.25 } & \red{\textbf{87.99*}} \\
          &  ~~~~  + DiverseCaption~\cite{Dunlap2023DiversifyYV} & 86.07  & \underline{95.44}  & \textbf{94.33} & \underline{87.12}  & \underline{92.01}  & 93.97  & 66.04  & \underline{87.85}  \\
          &  ~~~~  + InstructPix2Pix~\cite{Dunlap2023DiversifyYV} & 86.13  & 95.20  & 93.88  & 87.06  & 91.93  & 93.92  & 66.00  & 87.73  \\
\cmidrule{2-10}          & + \proposed{} \& MoEx & \underline{86.59}  & \underline{95.42}  & \underline{93.88}  & 86.34  & \underline{92.05}  & \textbf{94.19} & \textbf{67.51} & \underline{88.00}  \\
          & + \proposed{} \& RandAugment & \textbf{86.66} & \textbf{95.68} & \textbf{94.52} & \textbf{88.06} & \textbf{92.75} & \underline{94.13}  & \underline{67.45}  & \textbf{88.46} \\
          & + Text2Img \& \proposed{} & 85.87  & 95.16  & 93.76  & \underline{86.93}  & 92.04  & 93.74  & 64.63  & 87.45  \\
    \bottomrule
    \end{tabular}%
    }
    \caption{Accuracy of seven image classification datasets and three backbones by four baselines. On each backbone, the performance of the backbones, the perturbation-based methods, the generative methods (including \red{\proposed{}}), the integrated methods are provided. The best and second best results in the DA method and integrated methods for each dataset are \textbf{bolded} and \underline{underlined}. The numbers with * indicate that the im-provement of \proposed{} is statistically significant with $p < 0.05$ under t-test.}  
  \label{tab:main-results}%
\end{table*}%

\subsection{Image-To-Image Generation}\label{sec:Image-Generation}
In \textbf{Step 2}, we take the textual label $l$ and the caption ${c}^*$ as semantic guidance for image-to-image generation. 
We first concatenate $l$ and ${c}^*$ to construct the \textit{textual prompt}, \ie{} ${p} = ({l}, {c^*})$.
For example, ``A picture of a [Chevrolet Silverado 1500 Extended Cab 2012], a 2009 chevrolet silverado in a desert.'' in Figure~\ref{fig:main_method}.
The textual prompt carries not only accurate but brief image semantics from $l$, but also the overall image description from $c^*$.
It serves as a part of the input to Stable Diffusion to provide semantic guidance for augmented image generation.
As for image diversity, we apply Gaussian noise with a noise rate $n$ in Stable Diffusion to make slight changes to the original image based on the above semantic constraint.

Considering the different contributions of $l$ and ${c}^*$ to $p$,
we adopt the \textbf{prompt weighting} strategy.
Specifically, we apply different weights $w_l, w_c$ for the label and caption respectively by multiplying the token embeddings of $l$ and ${c}^*$ by $w_l$ and $w_c$ respectively.
Furthermore, to control the extent of semantic guidance on image generation, we adopt the \textbf{guidance mapping} strategy that provides a \textit{proper} guidance scale $g$. 
The guidance scale $g$ control how much the image generation process follows the semantic guidance, \ie{} the textual prompt $p$.
We apply a function $f$ to map the similarity $s^*$ between the original image ${x}$ and the chosen caption $c^*$, to the guidance scale $g= f({s}^*)$.
Finally, Stable Diffusion generates the augmented images given the above elements:

\begin{equation}
\label{eq:gen_image}
{x}' = \texttt{StableDiffusion}(x, p, g, n),
\end{equation}
where ${x}, {x}'$ is the original and augmented image, ${p}$ is the textual prompt, $g$ is the guidance scale, and $n$ is the noise rate.\looseness=-1

\section{Experiments}\label{sec:experiments}

\subsection{Experimental Settings}\label{sec:Datasets-Backbones}
\textbf{Datasets:} We evaluate the effectiveness of our proposed method on seven commonly used datasets, including three \textit{coarse-grained} object classification datasets: {CIFAR-10}, {CIFAR-100}~\cite{krizhevsky2009learning}, {Caltech101} (Cal101) ~\cite{DBLP:conf/cvpr/LiFP04}, and four \textit{fine-grained} object classification datasets: {Stanford Cars} (Cars)~\cite{krausecollecting}, {Flowers102} (Flowers)~\cite{nilsback2008automated}, {OxfordPets} (Pets)~\cite{parkhi2012cats} and texture classification DTD~\cite{cimpoi2014describing}. 

\textbf{Backbones:} We conduct experiments on three backbones, including a basic model from scratch: ResNet-50 (from scratch)~\cite{he2016deep}, and two pre-trained models: ViT (ImageNet-21k)~\cite{dosovitskiy2020image}, CLIP-ViT (LAION-2B)~\cite{Cherti2022ReproducibleSL}. 
Specifically, ViT (ImageNet-21k) is supervised trained on ImageNet-21k, while CLIP-ViT (LAION-2B) is self-supervised pre-trained on the CLIP paradigm on almost the same pre-trained corpus as the image generation model Stable Diffusion (SD).\footnote{For more details of CLIP-ViT (LAION-2B), see Appendix Sec.~\ref{sec:app-implementation}} 

\textbf{Baselines:} We apply various DA methods introduced in Sec.~\ref{sec:background} as baselines, including four perturbation-based methods:
\textit{Random Erasing} (RE), \textit{CutMix},  \textit{MoEx}, and \textit{RandAugment} (RA),
and three generative methods: \textit{Text2Img}, \textit{\proposed{}+DiverseCaption} (\textit{\proposed{}+DC}), and \textit{\proposed{}+InstructPix2Pix} (\textit{\proposed{}+IP}).
All generative methods employ the same image generation model SD.
we re-implement Text2Img, \proposed{}+DC, and \proposed{}+IP to provide extensive experiments across datasets and backbones.

\subsection{Implementation Details}
We apply nucleus sampling and beam search to respectively generate $10$ captions by BLIP.
The caption length is between $5$ and $20$.
We use $p = 0.9$ by default in nucleus sampling and $num\_beams$ = 3 by default in beam search.
The default ``CLIP-ViT-B/32'' model is used for calculating image-text similarity.
We apply the pre-trained ``stable-diffusion-v1-5'' model and generate one augmented image for each original image.~\footnote{SD v1-5 is mainly pre-trained on LAION-2B~\cite{Schuhmann2022LAION5BAO}.}
Empirically, we take $f(s^*) =-4 \cdot ({s^*}) ^{2}+ 2\cdot s^* + 1$ as the guidance mapping function.
We select the noise rate $n$ from $\{0.3, 0.5, 0.7\}$.\footnote{
The larger $n$ brings more variation.
We choose $n \in [0,1]$ from $\{0.3, 0.5, 0.7\}$ to 
preserve image semantics and bring explicit changes in the background, position, etc.}
As for prompt weighting, We assign a weight of $1.50$ to the labels because it carries more accurate information for the original image, and a weight of $0.90$ to the caption to reduce the interference caused by the potential low-quality captions. We run each method over 5 different random seeds.
See Appendix Sec.~\ref{sec:app-implementation} for more details including the training of image classifiers. \footnote{Our code will be released at \url{https://github.com/BohanLi0110/SGID}}

\subsection{Main Results}\label{sec:main-results}
In this paper, we conduct experiments on three backbones with seven strong DA baselines on seven datasets.
Our analysis of these results is based on three perspectives: \tif{(1)} overall performance on three backbones; \tif{(2)} average performance gain compared to the best baselines; \tif{(3)} average performance gain for combined models.
We believe the significant performance gain across the above backbones, datasets, and baselines demonstrates the effectiveness and generalizability of \proposed{}.
Our main results are shown in Table~\ref{tab:main-results} and Appendix Sec.~\ref{sec:appendix-3*3-histogram}. 

For \tif{(1)}, overall, \proposed{} shows positive effects and achieves the highest performance on average across all seven datasets and three backbones.
Specifically, our method leads to $10.39\%$ accuracy gains
on ResNet-50 (from scratch), $2.08\%$ on ViT (ImageNet-21k), and $0.85\%$ on CLIP-ViT (LAION-2B). 
This demonstrates that augmenting images with semantic guidance to diffusion models can benefit different backbones.
Notably, 
\proposed{} still shows improvement on CLIP-ViT (LAION-2B), whose pre-training data is almost identical to SD.
This demonstrates the effectiveness of the paradigm introduced in \proposed{}: preserving \textbf{semantic consistency} in original images and simultaneously bringing good \textbf{image diversity}, which will be further discussed in Sec.~\ref{sec:Consistency-Diversity}.

For \tif{(2)}, with semantic guidance, \proposed{} performs comparably or better than the best perturbation-based and generative baselines.
Specifically, \proposed{} outperforms \textit{RandAugment} and \textit{\proposed{}+DiverseCaption} by $4.89\%$ and $1.72\%$ on average on ResNet-50 (from scratch), by $1.26\%$ and $0.33\%$ on ViT (ImageNet-21k), and by $0.19\%$ and $0.14\%$ on CLIP-ViT (LAION-2B).
Higher performance than four strong baselines shows promising capabilities of \proposed{} to generate images with diffusion models under semantic guidance, \ie{} balancing diversity and semantic consistency.

For \tif{(3)}, \proposed{} can be combined with perturbation-based and generative baselines for further improvement.
We separately explore applying RandAugment based on \proposed{} and applying \proposed{} based on Text2Img.\footnote{Applying RandAugment based on \proposed{}'' indicates conducting the RandAugment transformation based on the augmented images of \proposed{}. ``Applying \proposed{} based on Text2Img'' indicates conducting \proposed{} based on the augmented images of Text2Img. We provide more results of combined models in Appendix Sec.~\ref{sec:plug-in-result}.}
We find the above integrated methods achieve further improvements, and this conclusion holds on three backbones.
For example, as a combination of ``perturbation-based \& generation-based'', ``\proposed{} \& RA'' exceeds RA on three backbones by $12.47\%$, $1.55\%$ and $0.66\%$, and exceeds \proposed{} by $7.58\%$, $0.29\%$ and $0.47\%$.
Consistent and significant performance gains further prove our \proposed{} not only preserves the essential semantics of the original images while bringing good diversity, but also benefits mutually with the perturbation-based method.
Interestingly, although ``generation-based'' methods, Text2Img and our \proposed{}, share the same image generation model, ``Text2Img \& \proposed{}'' still achieves performance gains compared with Text2Img ($2.39\%$, $0.25\%$ and $0.14\%$), but underperforms \proposed{} ($-0.77\%$, $-0.79\%$ and $-0.54\%$).
We attribute the gains to the semantic guidance introduced by our \proposed{}, but attribute the reductions to the fact that Text2Img may incorrectly change the essential semantics of original images.

\begin{table}[t]
  \centering

  \adjustbox{width=1.0\linewidth}{
    \begin{tabular}{lccccccccc}
    \toprule
          & \multicolumn{2}{c}{\textbf{Cal101}} & \multicolumn{2}{c}{\textbf{DTD}} & \multicolumn{2}{c}{\textbf{Pets}} & \multicolumn{2}{c}{\textbf{Avg.}} & {\textbf{Avg.}} \\
          & Con.  & Div.  & Con.  & Div.  & Con.  & Div.  & Con.  & Div.  &  {Performance} \\

    \midrule
    RandAugment & 4.49  & 1.30  & 4.51  & 1.50  & 4.70  & 1.34  & 4.57  & 1.38  & 63.03  \\
    \proposed{}   & 4.07  & 1.99  & 4.40  & 1.76  & 4.52  & 1.98  & 4.33  & 1.91  & \textbf{67.92}  \\
    \proposed{}+DC & 3.56  & 2.33  & 3.33  & 2.71  & 3.76  & 2.58  & 3.55  & 2.54  & \underline{66.20}  \\
    Text2Img & 1.67  & 4.56  & 1.27  & 4.87  & 1.41  & 4.62  & 1.45  & 4.68  & 64.76  \\
    \bottomrule
    \end{tabular}%
    }
    \caption{Human evaluation results of four DA methods on three datasets from the perspectives of semantic consistency (Con.) and Diversity (Div.). ``\proposed{}+DC'' indicates \proposed{}+DiverseCaption.}
  \label{tab:human-evaluation}%
\end{table}%

\subsection{Image Diversity and Semantic Consistency }\label{sec:Consistency-Diversity}

In this section, we aim to discuss the image diversity and semantic consistency  of \proposed{} and other baselines from three perspectives: \tif{(1)} human evaluation; \tif{(2)} automatic similarity evaluation; \tif{(3)} case study. We try to analyze the potential reason why \proposed{} achieves better performance than existing perturbation-based baselines and generative methods.

\subsubsection{Human Evaluation}\label{sec:human-evaluation}
We apply human evaluation on one coarse-grained object classification dataset (Caltech101), one fine-grained object classification dataset (OxfordPets), and the texture classification dataset (DTD).\footnote{The image size of CIFAR-10 and CIFAR-100 are $32*32$, which is too small for human evaluation.}
For each dataset, we randomly choose $10$ labels and $10$ of their corresponding original images.
We compare \proposed{} with the best perturbation-based and generative baselines: RandAugment, Text2Img, and \proposed{}+DiverseCaption.
We evaluate the augmented images by four DA methods based on the original image. Each augmented image is scored on a scale of $1\!\sim\!5$ in terms of image diversity and semantic consistency respectively.
We employ three experienced annotators. The annotators are trained and pass trial annotations. Each annotator spends an average of $4.5$ hours on annotation, and the salary for labeling each piece of data is \$$1$. 
Table~\ref{tab:human-evaluation} shows the human evaluation results for four methods and their corresponding average performance on ResNet-50 (from scratch).

\begin{figure}[!t]
  \centering 
  \includegraphics[width=0.95 \linewidth]{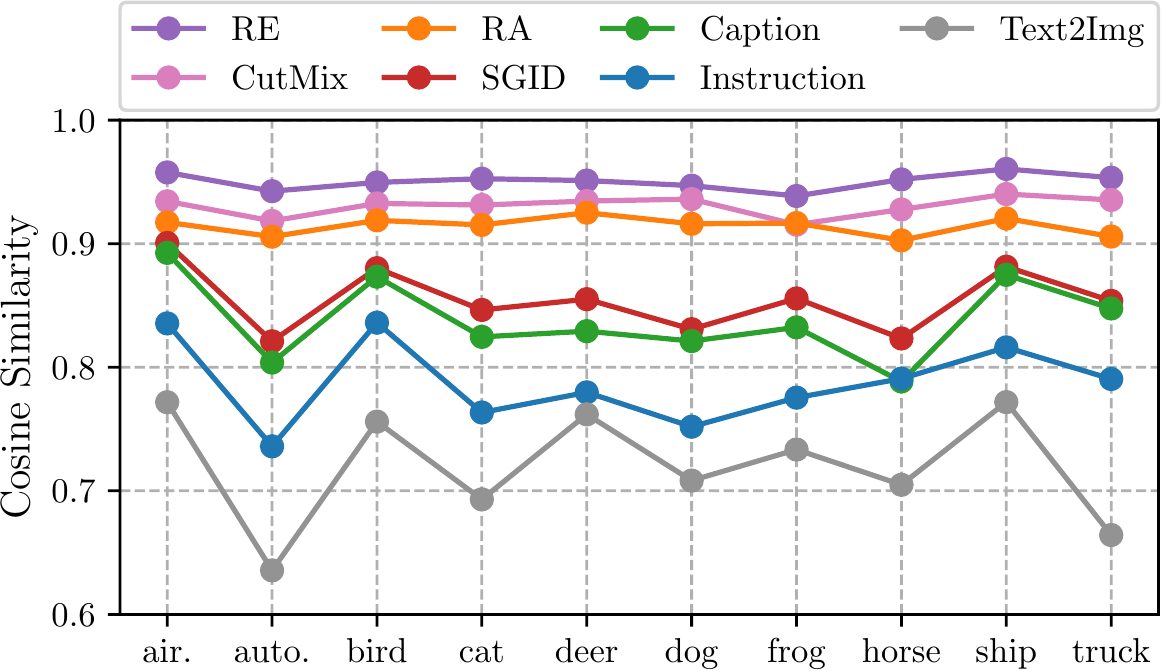}
  \caption{Average cosine similarity between the augmented and original images for each category of CIFAR-10 (air.: airplane, auto.: automobile).}
  \label{fig:similarity}
\end{figure}

We can find that \proposed{} has similar semantic consistency and more image diversity than RandAugment, but more semantic consistency and less image diversity than Text2Img and \proposed{}+DiverseCaption.
When both image classification performance and human evaluation results are considered, our \proposed{} achieves the best performance by balancing image diversity and semantic consistency through semantic-guided generative image augmentation.

\subsubsection{Automatic Similarity Evaluation}\label{sec:diverdity}
We choose CIFAR-10 for automatic similarity evaluation and separately use \proposed{} and six DA baselines to generate five augmented images for each original image.\footnote{We do not include MoEx since it performs the transformation in an implicit feature space instead of the explicit input space.}
For each DA method, we calculate the average cosine similarity between the original image and its five augmented ones~\cite{zhang2022expanding}.
We repeat this process on all original images and calculate the average value for each label as a measure of diversity. 
The lower the average similarity between the augmented images and the original image, the lower the semantic consistency but the more significant the diversity brought by the corresponding DA method. The results are shown in Figure~\ref{fig:similarity}.

\begin{figure}[!t]
  \centering 
  \includegraphics[width=1.0\linewidth]{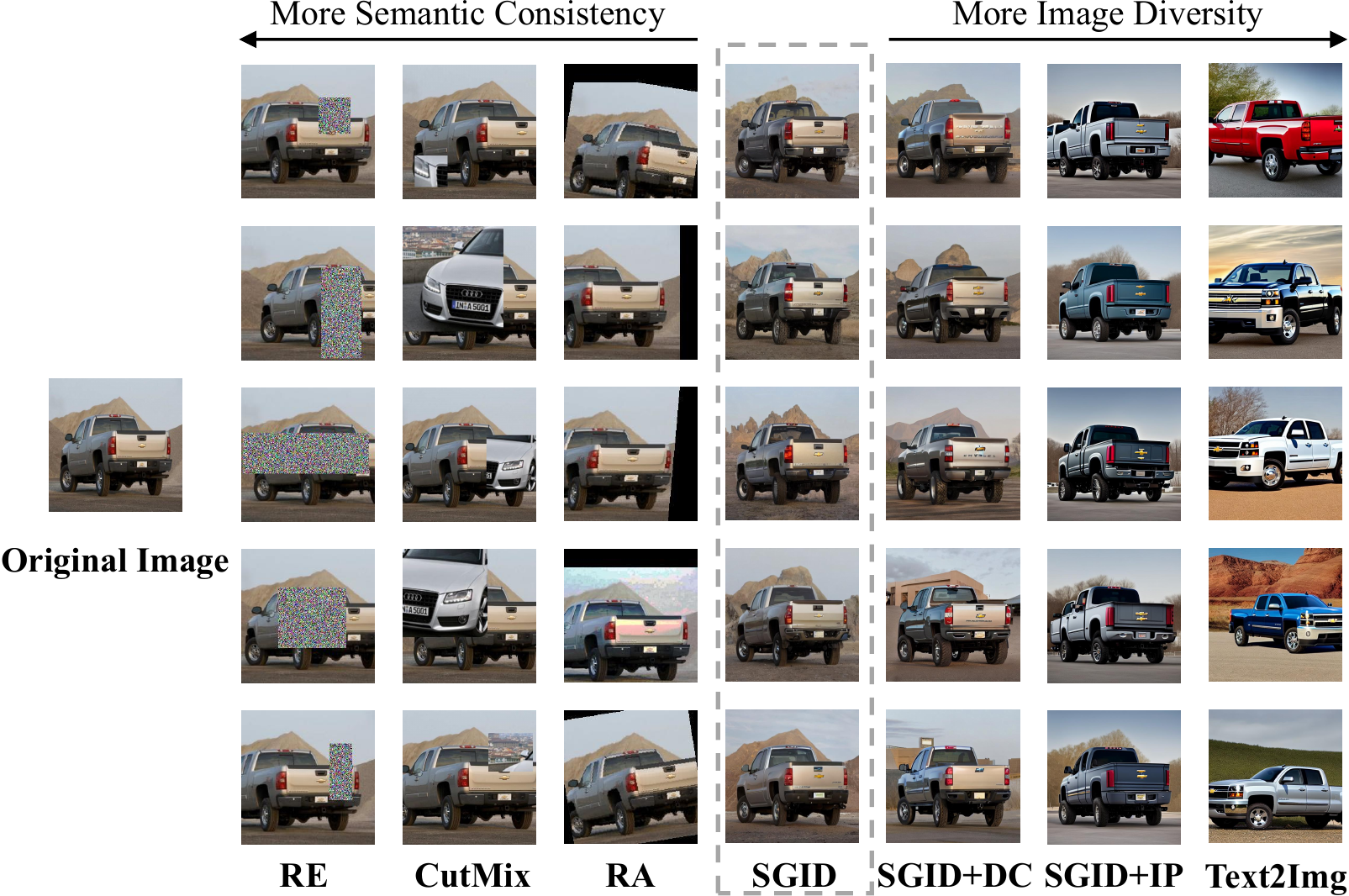}
  \caption{Case study of seven DA methods.
  }
  \label{fig:new-case-study_5X8}
\end{figure}

Overall, \proposed{} has a lower similarity ($0.8548$) compared to perturbation-based DA methods, while it has a higher similarity compared to other generative methods.
The average similarity of our \proposed{} is between the two categories of DA methods, but \proposed{} performs best in the image classification task.
This further demonstrates the significance of balancing image diversity and semantic consistency.

\subsubsection{Case Study}\label{sec:case-study}
Figure \ref{fig:new-case-study_5X8} compares the augmented images generated by \proposed{} and the other six baselines.\footnote{More case studies are in Appendix Figure~\ref{fig:appendix-case-study-cifar100} - Figure~\ref{fig:appendix-case-study-DTD}. 
}
The three perturbation-based methods bring diversity through transformations. 
However, these pre-defined transformations could not provide sufficient diversity for augmented images.
Generative baselines bring more diverse and vivid images than perturbation-based baselines, but they struggle to preserve the semantic consistency of original images. 
In contrast, \proposed{} preserves the semantic consistency from original images and provides good image diversity, which also results in a performance improvement in downstream tasks.

\subsection{Ablation Study}
In this section, we study the influence of some essential components in \proposed{}, including
\tif{(1)} the construction of the textual prompt;
\tif{(2)} the caption filter and the prompt weighting; 
\tif{(3)} the noise rate and the guidance scale.
We further explore augmented image filtering in Appendix Sec.~\ref{sec:image-filter}. 

\subsubsection{Influence of Textual Prompts}\label{sec:Textual-Prompt}

We study the influence of textual prompts based on ResNet-50 (from scratch) across four datasets in Table~\ref{tab:ablation}. We have the following conclusions:
\tif{(i)} {The semantic guidance is important for the image-to-image paradigm in \proposed{}.}
In most cases, \ie{} \textit{Cars}, \textit{Pets}, and \textit{DTD}, the performance of ``w/o Prompt'' is the lowest. 
This means that semantic guidance is essential for image augmentation to preserve semantic consistency.
\tif{(ii)} The label and the caption both provide semantic constraints.
In most cases except for \textit{Cars}, ``+ Complete Prompt'' shows the best results.
As for \textit{Cars}, ``+ Label Only'' has higher performance than ``+ Complete Prompt''. 
We attribute that the dataset is fine-grained and its labels carry very detailed information like ``Chevrolet\_Silverado\_1500\_Extended\_Cab\_2012'', which is sufficient for the PMs to generate an augmented image based on the original one. However, BLIP struggles to generate fine-grained captions, thus the generated captions have a counterproductive effect.
\tif{(iii)} {In most cases, ``+ Label Only'' outperforms ``+ Caption Only''} since the image label carries groundtruth and accurate image semantics.
However, for \textit{CIFAR-100}, ``+ Caption Only'' shows a better result than ``+ Label Only''.
We attribute to the short label, \eg{} bed, forest, etc., provided by \textit{CIFAR-100} cannot provide detailed information in the label like other fine-grained datasets.

\begin{table}
\small
  \centering
  \adjustbox{width=0.8\linewidth}{
\begin{tabular}{lcccc}
    \toprule
\textbf{}   & \textbf{CIFAR-100} & \textbf{Cars}  & \textbf{Pets}   & \textbf{DTD}  \\ \midrule
w/o Prompt  & 74.11              &  82.09         &  55.76         & 32.39          \\ 
+ Caption Only  & \underline{75.65}        & 83.89          &   63.80        &  34.78         \\
+ Label Only & 73.99              &   \textbf{88.53}  & \underline{75.90}    & \underline{37.43} \\
+ Complete Prompt      & \textbf{75.72}     & \underline{84.31} & \textbf{76.15} &  \textbf{37.55}   \\
    \bottomrule
\end{tabular}
}
     \caption{
     Ablation study of textual prompts. ``w/o Prompt'' indicates no semantic guidance, ``+ Caption Only'', ``+ Label Only'', and ``Complete Prompt'' use the caption only, the label only, and the complete prompt as the guidance.
     }
  \label{tab:ablation}%
\end{table}

\begin{table}
\small
  \centering
    \adjustbox{width=0.9\linewidth}{

\begin{tabular}{lcccccc}
\toprule
\multirow{2}{*}{} & \multicolumn{2}{c}{\textbf{CIFAR-100}} & \multicolumn{2}{c}{\textbf{Pets}} & \multicolumn{2}{c}{\textbf{DTD}} \\
                  & PW                  & w/o PW        &  PW       & w/o PW              &  PW               & w/o PW     \\ \midrule
Beam              & 75.60                   & 74.69         & 75.06       & \textbf{76.15}      & 37.47               & 35.79      \\
Nucleus           & \textbf{75.72}         & 74.93         & 73.72       & 75.70       & 37.16               & 37.04      \\
Caption Filter              & 74.80                   & 74.85         & 70.86       & 71.95               & \textbf{37.55}      & 32.44      \\ \bottomrule
\end{tabular}
}
     \caption{
     Ablation study of caption sampling and prompt weighting. 
    ``Caption Filter'' indicates using CLIP to choose the caption with the highest similarity to the original image among the $20$ captions generated by ``Beam'' and ``Nucleus''.
    ``PW'' indicates the prompt weighting strategy.
     }
  \label{tab:analysis}
\end{table}

\subsubsection{Caption Sampling \& Prompt Weighting}
We study the influence of caption sampling and prompt weighting based on ResNet-50 (from scratch) in Table~\ref{tab:analysis}. 
Nucleus sampling, beam search, and the caption filter contribute the best performance on the three datasets respectively. 
This may be because of the different characteristics of nucleus sampling and beam search, as mentioned in Section~\ref{sec:Image-Generation}.
Nucleus sampling generates more diverse and surprising captions, while beam search tends to generate safe captions~\cite{li2022blip}.
We hypothesize that \textit{Pets} is a fine-grained dataset whose labels are very detailed and a ``safe'' caption would not influence the semantics of the labels.
In contrast, short labels in \textit{CIFAR-100} cannot convey sufficient semantics, thus more diverse captions from nucleus sampling are required.
\textit{DTD}, the texture classification dataset, is challenging for BLIP to generate captions ensuring image semantics.
Then the caption filter would help to improve caption quality for semantic consistency.
As for prompt weighting, its effectiveness is proven in two out of three datasets.
We hypothesize that adding different weights to the label and the caption could reconcile their different semantic information.

\subsubsection{Influence of Noise Rate and Guidance Scale.}

\begin{figure}[!t]
    
  \centering 
  \includegraphics[width=0.8\linewidth]{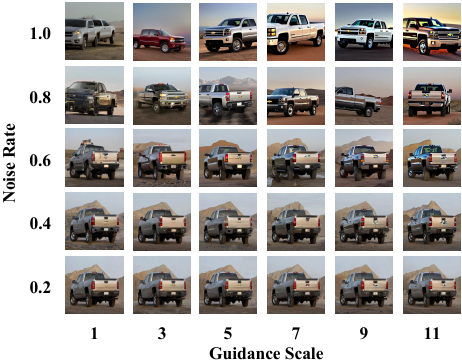}
  \caption{
  Case study on the influence of noise rate and guidance scale when generating images by \proposed{} on Cars.
}
  \label{fig:case-study}
\end{figure}

We further explore the effect of adjusting the noise rate and guidance scale on generating augmented images.
As shown in Figure \ref{fig:case-study}(b), the generated images show more diversity (\eg{} variations in body orientation, color, background, etc.) when the noise rate and the guidance scale are increased.
However, the experimental results show that as the noise rate increases, the performance decreases while more diversity is introduced.
This suggests that noise rate has a great effect on performance and deserves further exploration in future work.

\section{Related Works}
\textbf{Diffusion Model-Based Methods.}
There are some works employing diffusion models to generate augmented images or expand datasets.
\citet{zhang2022expanding}, \citet{He2022IsSD}, and \citet{Dunlap2023DiversifyYV} generate images based on label-related constraints and/or original images.
\citet{Li2023YourDM} and \citet{Shipard2023DiversityID} generate samples for training zero-shot classifiers without relying on original images.
While acknowledging the importance of zero-shot learning, this paper focuses on linear probing to improve model performance in image classification.
\citet{Akrout2023DiffusionbasedDA}, \citet{Carlini2023ExtractingTD}, and \citet{Bakhtiarnia2023PromptMixTD} apply diffusion models for DA in other tasks like skin disease classification, privacy attacks, and crowd counting. These are promising directions we would explore in the future.

\paragraph{Knowledge Distillation (KD).} 
There have been some works using PMs to generate training data~\cite{meng2022generating,wang2021towards} and referring to it as a variant of KD~\cite{ye2022zerogen}.
To some extent, our work could also be categorized as KD. 
Semantic-guided image generation via pre-trained diffusion models can be seen as a more straightforward and effective form to precisely distill knowledge from pre-trained diffusion models.
Moreover, it can be naturally combined with the traditional KD as a new direction.

\section{Conclusion}
In this paper, we introduce \textbf{\proposed{}}, a \textbf{S}emantic-guided \textbf{G}enerative \textbf{I}mage augmentation method with \textbf{D}iffusion models for image classification.
The proposed method balances image diversity and semantic consistency.
Specifically, \proposed{} constructs prompts with image labels and captions as semantic guidance to generate augmented images that preserve the essential semantics in original images and simultaneously bring good image diversity.
We demonstrate the effectiveness of \proposed{} by experiments on three backbones with seven strong image augmentation baselines on seven different datasets and \proposed{} outperforms the backbones on all datasets and achieves the best or comparable performance to all baselines. 
Moreover, \proposed{} can be combined with other image augmentation baselines and further improves the overall performance. 
We also evaluate the semantic consistency and image diversity of \proposed{} through quantitative human and automated evaluations, as well as qualitative case studies.

\section*{Acknowledgments}
We gratefully acknowledge the support of the National Natural Science Foundation of China (NSFC) via grant 62236004 and 62206078, and the support of Du Xiaoman (Beijing) Science Technology Co., Ltd.

\bibliography{aaai24}

\appendix

\clearpage

\section{Combined Experiments with CutMix} \label{sec:plug-in-result}
As mentioned in Sec. 4.3, \proposed{} can be combined with other image augmentation baselines and further improves the overall performance.
We do not conduct combined experiments for \proposed{}+DiverseCaption or \proposed{}+InstructPix2Pix since they are the variants of \proposed{}.
We introduce the combined experimental results between \proposed{} and CutMix in this section, as shown in Appendix Table~\ref{tab:appen-main-results}.

We explore applying CutMix based on \proposed{}, \ie{} conducting the CutMix transformation based on the augmented images of \proposed{}.
We find the combined method achieves the most obvious performance gain on Resnet-50 (from scratch). Specifically, “\proposed{} \& CutMix” exceeds both CutMix and \proposed{} by 11.25\% and 1.99\%.
The significant performance gains further prove that our \proposed{} brings good diversity by diffusion models and preserves the essential original semantics by semantic guidance. This paradigm still benefits the perturbation-based CutMix on ResNet-50 (from scratch). As for Vit (ImageNet-21k) and CLIP-ViT (LAION-2B), “\proposed{} \& CutMix” achieves slightly better or comparable results to CutMix, and we will explore the reason in the future.

\section{Image Filters}\label{sec:image-filter}
We explore the effects of image filters for potential better performance. Three categories of image filters are conducted. \tif{(1)} Label Filter: we use CLIP~\cite{radford2021learning} to predict whether the augmented image is related to the groundtruth label or not. For example, in \textit{CIFAR-100}, we provide the list of textual labels including “a photo of an airplane”, “a photo of a bird”, “a photo of a horse”, ..., “a photo of a truck”. All images under the \textit{bird} label that are not classified as “a photo of a bird” are removed. \tif{(2)} Prompt Filter: we use CLIP to calculate the similarity between each single augmented image and the prompt used to generate this augmented image. The average similarity under each label is calculated and we apply it as the threshold. The augmented images with similarities lower than the threshold will be removed. Thus, images of some proportions will be removed in each label. \tif{(3)} Original Image Filter: we calculate the similarity between each single augmented image and the original image. The average similarity under each label is calculated and we apply it as the threshold. The augmented images with similarities lower than the threshold will be removed. Thus, images of some proportions will be removed in each label. We conduct this experiment for analysis and we have no objection to potential further improvements with more carefully designed filters.

We conduct experiments on ResNet-50 (from scratch), as shown in Appendix Table~\ref{tab:appen-main-results}. 
The \textit{Label Filter} outperforms \proposed{} on CIFAR-100 and CIFAR-10, which we attribute that the datasets are coarse-grained and the image size is very small, thus a filter tends to improve the performance. 
The \textit{Prompt Filter} outperforms \proposed{} on Pets and DTD, which we attribute that the generated captions bring more vivid and detailed semantic information than the labels, thus the prompt filter performs the best. 
The \textit{Original Image Filter} does not achieve performance gain on each dataset, which we attribute that the images filtered out through the original image are diverse ones that have a positive effect on image classification. That is, our \proposed{} tends to have a very limited amount of  low-level failure.
\proposed{} achieves the best performance on average. We attribute that our method introduces semantic guidance to preserve the semantic consistency from the original images, and this confirms the quality of the augmented image.

\section{Analysis on Two Strong Baselines}\label{sec:appendix-3*3-histogram}
\begin{figure*}[!t]
  \centering 
  \includegraphics[width=1.0\linewidth]{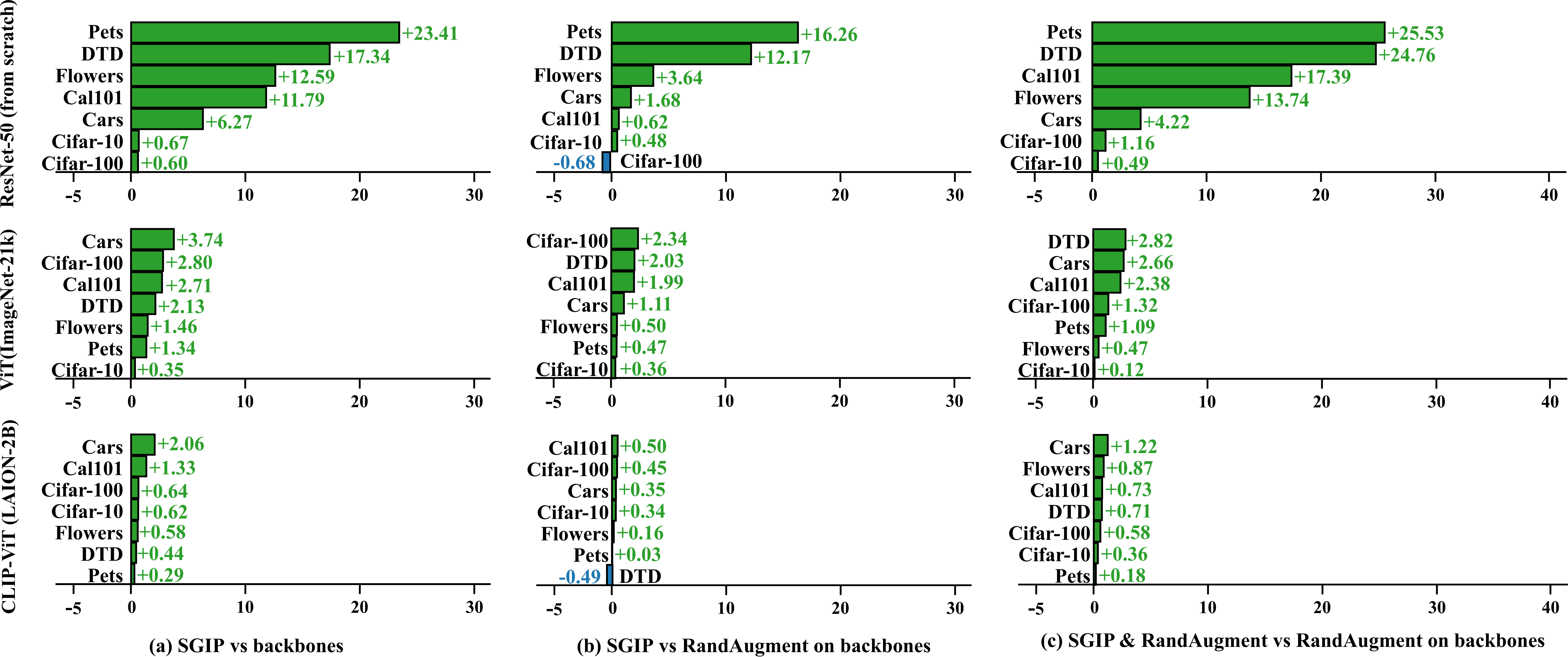}
  \caption{\proposed{} outperforms not only all backbones but also the generative baseline Text2Img. Moreover, \proposed{} can further improve the performance of Text2Img by combining naturally with it.}
  \label{fig:RandomAugment}
\end{figure*}

\begin{figure*}[!t]
  \centering 
  \includegraphics[width=1.0\linewidth]{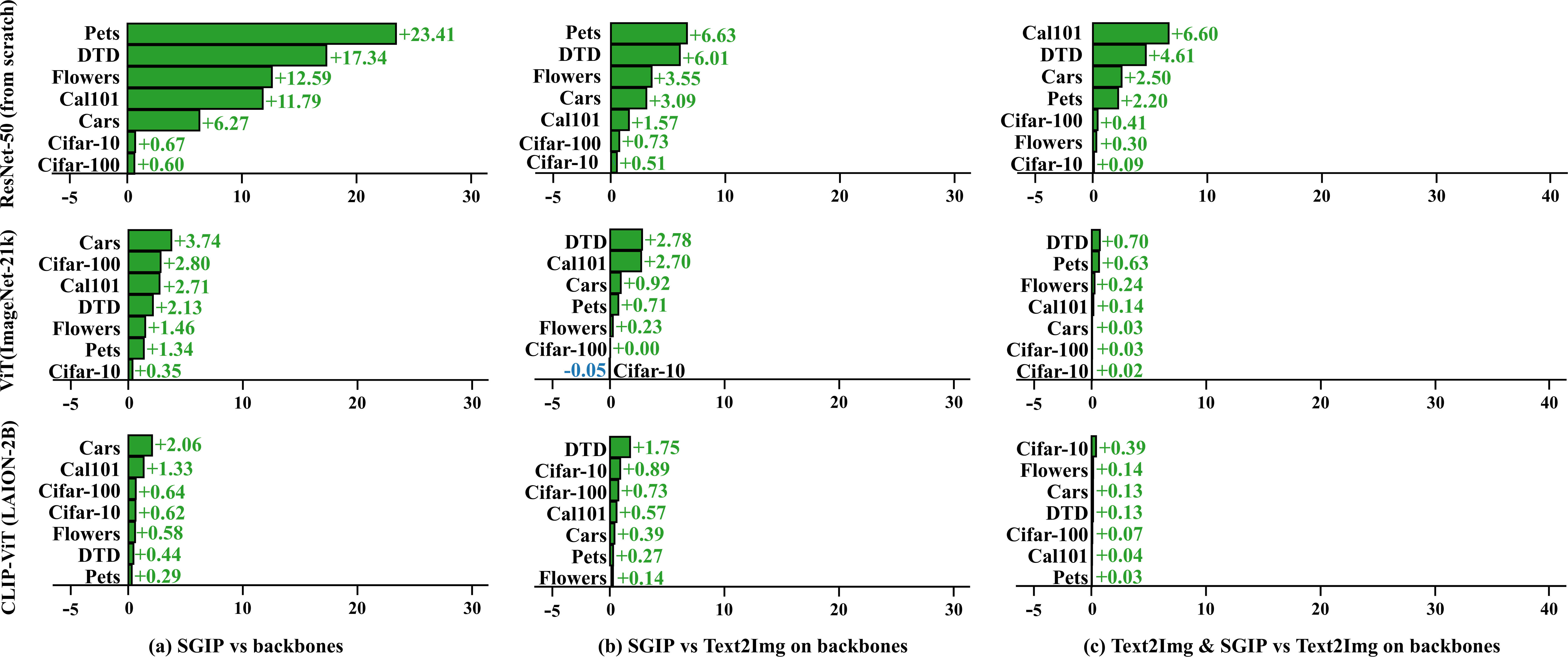}
  \caption{\proposed{} outperforms not only all backbones but also the best perturbation-based baseline RandAugment. Moreover, \proposed{} can further improve the performance of RandAugment by combining naturally with it.}
  \label{fig:Text2Img}
\end{figure*}

To show the effectiveness of \proposed{}, we provide a detailed analysis with two strong baselines across three backbones from three perspectives. We choose the perturbation-based RandAugment as well as the generative Text2Img as our focus. The results are shown in Fig.~\ref{fig:RandomAugment}-\ref{fig:Text2Img}.
\tif{(1)} \proposed{} vs the backbones: \proposed{} surpasses all three backbones on all datasets, specifically, the performance gain on ResNet-50 (from scratch) is the most obvious, and we still see improvement on CLIP-VIT (LAION-2B), whose pre-training data is almost identical to Stable Diffusion. This shows the effectiveness of the paradigm of \proposed{}.
\tif{(2)} \proposed{} vs the strong baselines: Our method outperforms two strong baseline models on the vast majority of all datasets. This conclusion is valid on all three backbones. This is because \proposed{} brings good diversity by diffusion models and preserves semantic consistency through semantic guidance. It balances image diversity and semantic consistency, and archives better performance than strong perturbation-based and generative baselines.
\tif{(3)} \proposed{} can be combined with other baselines for further performance improvement, and this conclusion holds for both perturbation-based baselines and generative baselines. For example, both \textit{\proposed{} \& RandAugment} and \textit{Text2Img \& \proposed{}} outperforms the baselines themselves on all three backbones. Consistent and significant performance gains further prove our \proposed{} not only preserves the essential semantics of the original images while bringing good diversity, but also benefits the baselines.

\begin{table}
  \centering
  \setlength{\tabcolsep}{4pt}
  \adjustbox{width=0.9\linewidth}{
  \begin{tabular}{lccccc}
    \toprule
            & \textbf{CIFAR-100-S} & \textbf{Cars}   & \textbf{Pets}  & \textbf{DTD}   \\
    \midrule
     Backbone & 39.90  & 82.15 & 49.19 & 14.57  \\
           + GIF & 61.10 ($\times$ 5)  & 75.70 ($\times$ 20)  &  73.40 ($\times$ 30)  & 43.40 ($\times$ 20)  \\
           + \proposed{} & \textbf{65.20} ($\times$ 2) & \textbf{88.28} ($\times$ 1) & \textbf{76.15} ($\times$ 1) & \textbf{45.12} ($\times$ 2)  \\
    \bottomrule
    \end{tabular}%
  }       
    \caption{
    Comparison between GIF and \proposed{}.
    ($\times$ $n$) indicates that the augmented data is $n$ times the original data. 
    }
  \label{tab:appendix-expand-results}%
\end{table}%

\section{Comparison with Image Variation}\label{sec:appendix-gif}

We compare our approach with GIF~\cite{zhang2022expanding}, an image variation method conditioned on CLIP image embeddings.\footnote{We introduce it in Appendix Sec.~\ref{sec:appendix-generative-method}}
They perturb and optimize the latent features of the original image and input them to a pre-trained image generation model to generate the image.
However, they use the original images only as the input. The image labels are only used during optimizing the training loss, instead of used for constructing the textual prompt along with image captions as input, \ie{} as a \textbf{explicit} semantic guidance for image-to-image generation in \proposed{}.
Moreover, their methods require additional training on the pre-trained image generation models, and more times of expansion (usually 5-30 times) are required to achieve a boost on the datasets.

\begin{table*}[!ht]
  \centering
  \adjustbox{width=0.95\linewidth}{
    \begin{tabular}{clcccccccc}
    \toprule
          &       & \textbf{CIFAR-100} & \textbf{CIFAR-10} & \textbf{Cal101} & \textbf{Cars} & \textbf{Flowers} & \textbf{Pets} & \textbf{DTD} & \textbf{Avg.} \\
    \midrule
    \multirow{4}[3]{*}{ResNet-50 (from scratch)} & Backbone~\cite{he2016deep} & 75.12  & 94.81  & 47.38  & 82.26  & 33.02  & 50.30  & 19.85  & 57.53  \\
\cmidrule{2-10}          
          & + CutMix~\cite{yun2019cutmix} & \underline{77.56} & \underline{95.52} & 49.28  & 83.55  & 33.35  & 51.21  & 20.14  & 58.66  \\
          & {+ \proposed{} (Ours)} & {75.72} & {95.48} & {\underline{59.17}} & {\underline{88.53}} & {\underline{45.61}} & {\underline{73.71}} & {\underline{37.19}} & {\underline{67.92}} \\
          & + \proposed{} \& CutMix & \textbf{77.81}  & \textbf{95.74}  & \textbf{61.75} & \textbf{90.50} & \textbf{46.57} & \textbf{75.82} & \textbf{41.15} & \textbf{69.91} \\
    \midrule
    \multirow{4}[3]{*}{ViT (ImageNet-21k)} & Backbone~\cite{dosovitskiy2020image} & 89.86  & 98.61  & 91.20  & 82.99  & 95.70  & 92.04  & 71.22  & 88.80  \\
\cmidrule{2-10}          
          & + CutMix~\cite{yun2019cutmix} & 89.94  & 98.67  & 91.77  & 84.20  & 96.53  & 92.45  & 71.32  & 89.27  \\
        & {+  \proposed{} (Ours)} & {\textbf{92.66}} & {\textbf{98.96}} & {\underline{93.91}} & {\underline{86.73}} & {\underline{97.16 }} & {\textbf{93.38}} & {\textbf{73.35}} & {\underline{90.88}} \\
        
          & + \proposed{} \& CutMix & \underline{91.86}  & \underline{98.85}  & \textbf{94.20} & \textbf{87.45} & \textbf{97.66}  & \underline{92.94} & \underline{73.29} & \textbf{90.89} \\
    \midrule
    \multirow{4}[4]{*}{CLIP-ViT (LAION-2B)} & Backbone~\cite{Cherti2022ReproducibleSL} & 85.89  & 95.04  & 92.96  & 85.13  & 91.46  & 93.69  & 65.81  & 87.14  \\
\cmidrule{2-10}          
          & + CutMix~\cite{yun2019cutmix} & 85.98  & 95.20  & 93.85  & 85.43  & 91.64  & \underline{94.01} & 66.20  & 87.47  \\
          & {+ \proposed{} (Ours)} & {\textbf{86.53}} & {\textbf{95.66}} & {\underline{94.29}} & {\textbf{87.19}} & {\underline{92.04}} & {\underline{93.98}} & {\underline{66.25}} & {\underline{87.99}} \\

          & + \proposed{} \& CutMix & \underline{86.19} & \underline{95.34} & \textbf{94.46} & \underline{86.49} & \textbf{92.33} & \textbf{94.55}  & \textbf{66.73}  & \textbf{88.01} \\
    \bottomrule
    \end{tabular}%
    }
    \caption{The results of the combined model with CutMix on three backbones, \ie{} conducting the CutMix transformation based on the augmented images of \proposed{}.}  
  \label{tab:appen-main-results}%
\end{table*}%

\begin{table*}[!ht]
  \centering
  \adjustbox{width=0.95\linewidth}{
    \begin{tabular}{clcccccccc}
    \toprule
          &       & \textbf{CIFAR-100} & \textbf{CIFAR-10} & \textbf{Cal101} & \textbf{Cars} & \textbf{Flowers} & \textbf{Pets} & \textbf{DTD} & \textbf{Avg.} \\
    \midrule
    \multirow{3}[8]{*}{ResNet-50 (from scratch)} & Backbone~\cite{he2016deep} & 75.12  & 94.81  & 47.38  & 82.26  & 33.02  & 50.30  & 19.85  & 57.53  \\
\cmidrule{2-10}          
          & {+ \proposed{} (Ours)} & {\underline{75.72}} & {\underline{95.48}} & {\textbf{59.17}} & {\textbf{88.53}} & {\textbf{45.61}} & {73.71} & {\underline{37.19}} & {\textbf{67.92}} \\
          & ~~~~+ Label Filter & \textbf{76.23}  & \textbf{95.93}  & \underline{58.69} & \underline{87.56} & \underline{44.22} & \underline{75.01} & {32.71} & {67.19} \\
          & ~~~~+ Prompt Filter & {75.29}  & 95.44  & {57.37} & 86.02 & {43.75} & \textbf{75.47} & \textbf{38.99} & \underline{67.48} \\
          & ~~~~+ Original Image Filter & 75.16  & 95.26  & {53.69} & 86.76 & {40.88} & {68.33} & {36.97} & {65.29} \\
    \bottomrule
    \end{tabular}%
    }
    \caption{The performance of three categories of image filters. }  
  \label{tab:appen-main-results}%
\end{table*}%

We provide a fair comparison with GIF based on Stable Diffusion and ResNet-50 (from scratch) in Appendix Table~\ref{tab:appendix-expand-results}.
Since GIF does not provide their code or models, in Appendix Table~\ref{tab:appendix-expand-results}, we provide the results of ``Backbone'' from our experiments. ``CIFAR-100-S'' indicates 100-shot of CIFAR-100.
Experiments show that GIF underperforms our proposed \proposed{}, which we attribute to the performance fluctuations of CLIP image embedding on different datasets~\cite{radford2021learning} and the optimization difficulty of GIF.
Unlike optimizing the image features with label constraints, \proposed{} proposes to construct prompts with golden labels and captions to better guide the image-to-image generation process.

This provides more precise and explicit semantic information than the implicit semantic constraints via optimizing losses based on labels. 
Significant improvements with a smaller number of dataset expansion multiples (up to 2 times on CIFAR-100-S) prove the effectiveness of \proposed{}.
Besides, our method is also applicable to larger datasets (such as CIFAR-10 and CIFAR-100) and require no additional training on the pre-trained image generation models.
The results also show that \proposed{} preserves semantic consistency by semantic guidance, which is important for image classification. Our \proposed{} obtains obvious performance gain by balancing image diversity and semantic consistency.

\section{Image-Text Pre-trained Models}\label{sec:appendix-PMS}
Benefiting from vision-language pre-training on large-scale image-text pairs~\cite{jia2021scaling,changpinyo2021conceptual}, there has been a lot of work successfully unifying text and image in the united space~\cite{cho2021unifying,wang2021simvlm,xu2022bridge}.

Overall, there are three kinds of vision-language pre-trained models used in this paper:
\tif{(1)} \textbf{CLIP}~\cite{radford2021learning} consists of a text encoder and an image encoder that encodes both images and texts and successfully bridges the feature space between language and vision. We use it to measure the similarity between images and texts.
\tif{(2)} \textbf{BLIP}~\cite{li2022blip}, an encoder-decoder model jointly pre-trained with three vision-language objectives: image-text contrastive learning, image-text matching, and image-conditioned language modeling, achieves great results on a wide range of vision-language tasks such as image captioning and visual question answering.
\tif{(3)} \textbf{Stable Diffusion}~\cite{rombach2022high}, an image generation model pre-trained on LAION-2B~\cite{Schuhmann2022LAION5BAO}, captures the essential image information by compressing images from the pixel space to the latent space, and gradually adding Gaussian noise for the diffusion process. The CLIP text encoder converts the text description into the condition for the denoising process, and the image decoder decodes the denoised image into the final image. 

In this paper, \textbf{the reason why we choose Stable Diffusion as the image generation model} is that it is a widely used and highly effective open-source \textit{image-to-image} generation model that supports \textit{semantic injection} and noise rate tuning, which perfectly matches our approach (see Sec.~3). Moreover, the diffusion and conditional denoising process provides \textit{good diversity} for image generation and achieves good results on some downstream tasks including image generation. The significant performance of \proposed{} also justifies the choice of Stable Diffusion (see Sec.~4).

\begin{table*}[]
\small
  \centering
\begin{tabular}{lc|ccccc}
\toprule
\multirow{2}{*}{\textbf{Dataset}} & \multirow{2}{*}{\textbf{Image Label Example}}                 & \multirow{2}{*}{\textbf{Label Guidance}} & \multicolumn{3}{c}{\textbf{Caption Guidance}} & \multirow{2}{*}{\textbf{Prompt Weighting}} \\ \cline{4-6}
                                  &                                                               &                                          & \textbf{Beam}         & \textbf{Nucleus}         & \textbf{CLIP}         &                                            \\ \hline
CIFAR-100                         & forest, fox, bed                                              & \ding{52}                                        & \ding{56}             & \ding{52}               &\ding{56}              & \ding{52}                                         \\
CIFAR-10                          & airplane, bird, ship                                          & \ding{52}                                         &\ding{56}              &\ding{56}                 & \ding{52}             & \ding{52}                                           \\
Cal101                            & anchor, cellphone, pyramid                                     & \ding{52}                                         &\ding{56}              & \ding{52}               &\ding{56}              &\ding{56}                                            \\
Cars                              & \begin{tabular}[c]{@{}c@{}}Acura\_Integra\_Type\_R\_2001, \\ AM\_General\_Hummer\_SUV\_2000\end{tabular} & \ding{52}                                        &\ding{56}              & \ding{56}                &\ding{56}              &    \ding{56}                                        \\
Flowers                           & moon orchid, sword lily                                       & \ding{52}                                         &   \ding{56}           &    \ding{56}             & \ding{52}            & \ding{52}                                           \\
Pets                              & abyssinian, shiba inu                                         & \ding{52}                                         & \ding{52}            &   \ding{56}              &    \ding{56}          &    \ding{56}                                        \\
DTD                               & banded, zigzagged                                             & \ding{52}                                        &    \ding{56}          &   \ding{56}              & \ding{52}             & \ding{52}                                          \\ \bottomrule
\end{tabular}
    \caption{Detailed experimental settings on label guidance, caption guidance and prompt weighting.}
  \label{tab:dataset_setting}
\end{table*}

\section{Case Study of DA Methods}
\begin{figure*}[!t]
  \centering 
  \includegraphics[width=0.8 \linewidth]{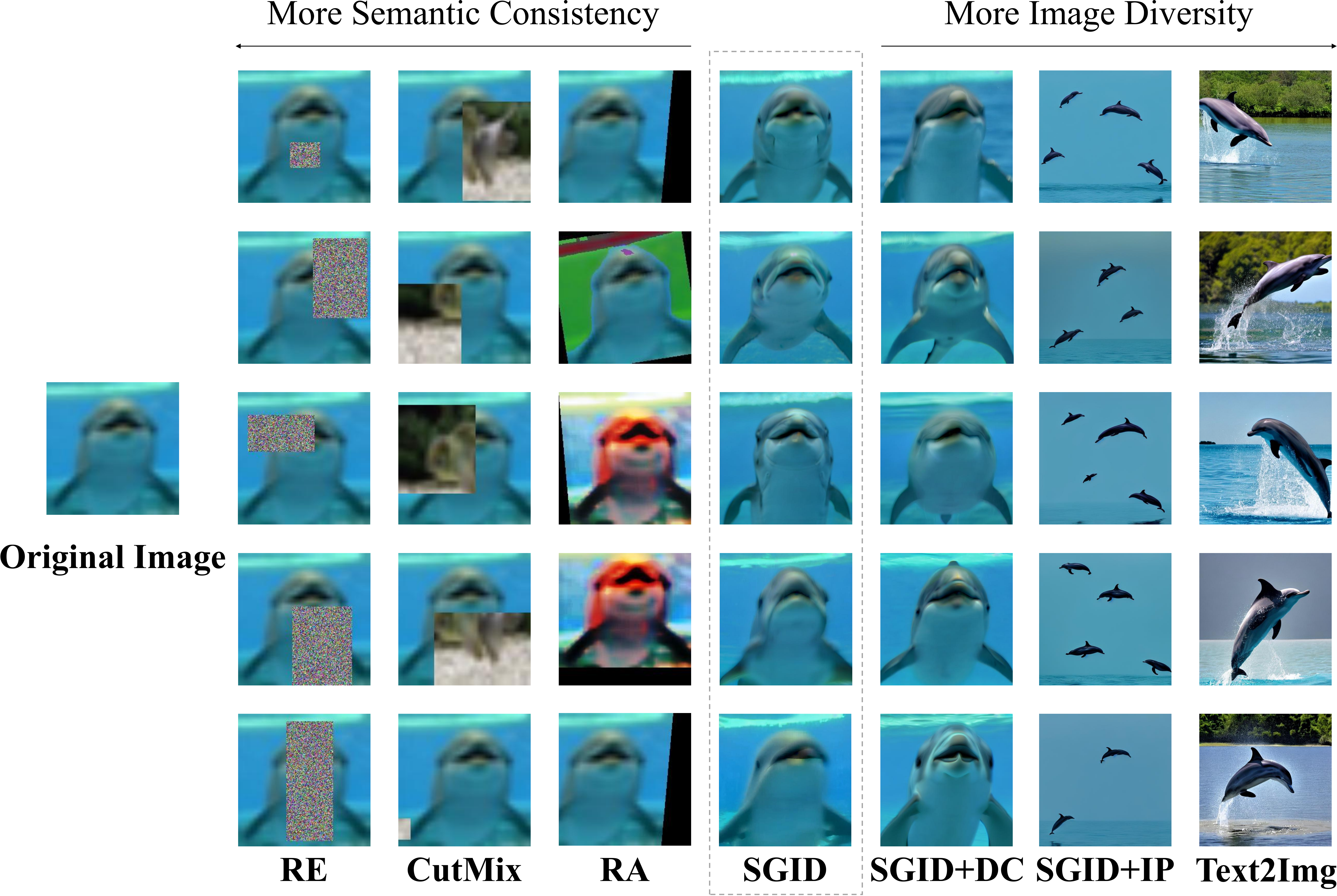}
  \caption{Augmented images by \proposed{} and other DA baselines on CIFAR-100.}
  \label{fig:appendix-case-study-cifar100}
\end{figure*}

\begin{figure*}[!t]
  \centering 
  \includegraphics[width=0.8 \linewidth]{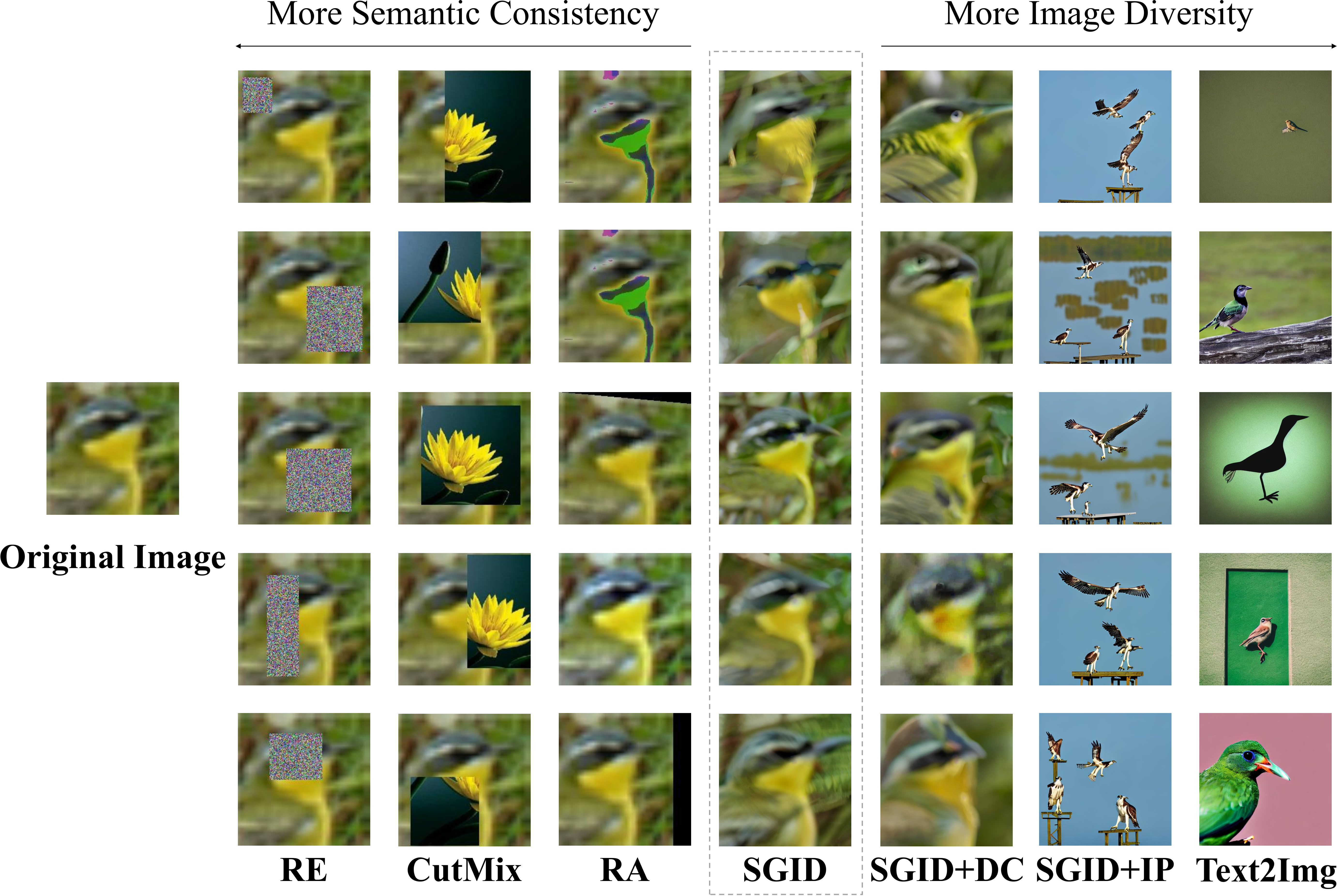}
  \caption{Augmented images by \proposed{} and other DA baselines on CIFAR-10.}
  \label{fig:appendix-case-study-cifar10}
\end{figure*}

\begin{figure*}[!t]
  \centering 
  \includegraphics[width=0.8 \linewidth]{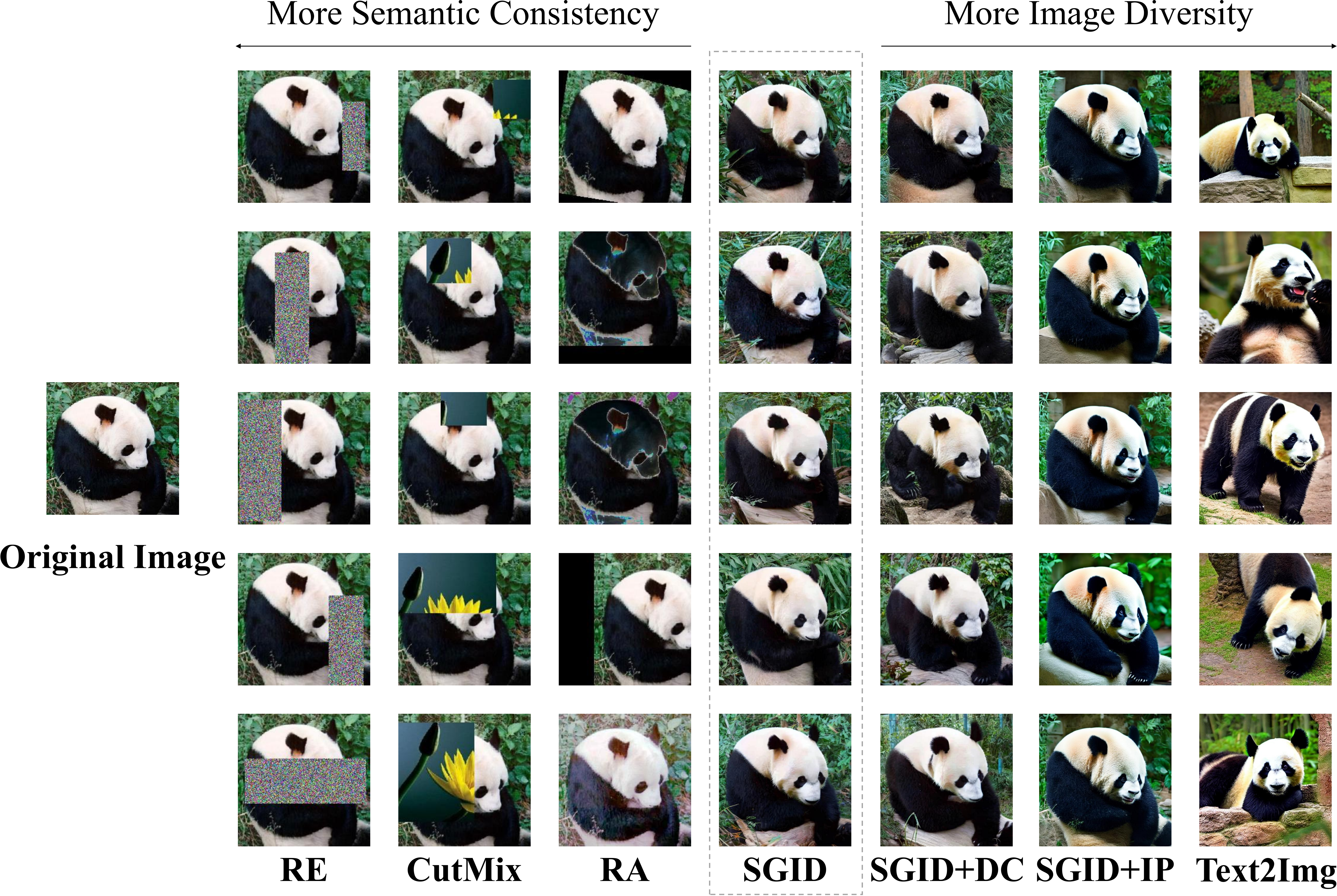}
  \caption{Augmented images by \proposed{} and other DA baselines on Caltech101.}
  \label{fig:appendix-case-study-cal101}
\end{figure*}

\begin{figure*}[!t]
  \centering 
  \includegraphics[width=0.8 \linewidth]{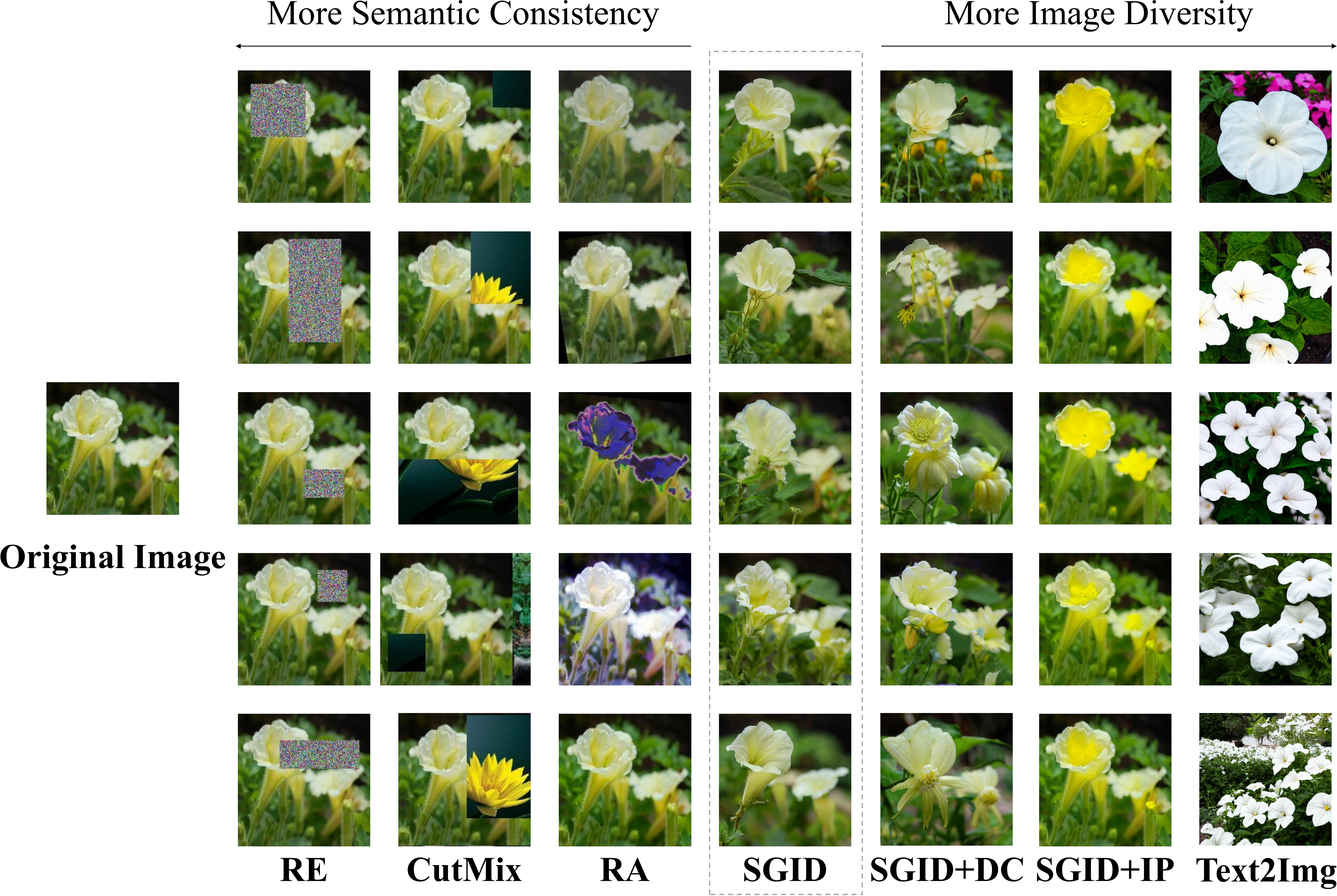}
  \caption{Augmented images by \proposed{} and other DA baselines on Flowers102.}
  \label{fig:appendix-case-study-flower}
\end{figure*}

\begin{figure*}[!t]
  \centering 
  \includegraphics[width=0.8 \linewidth]{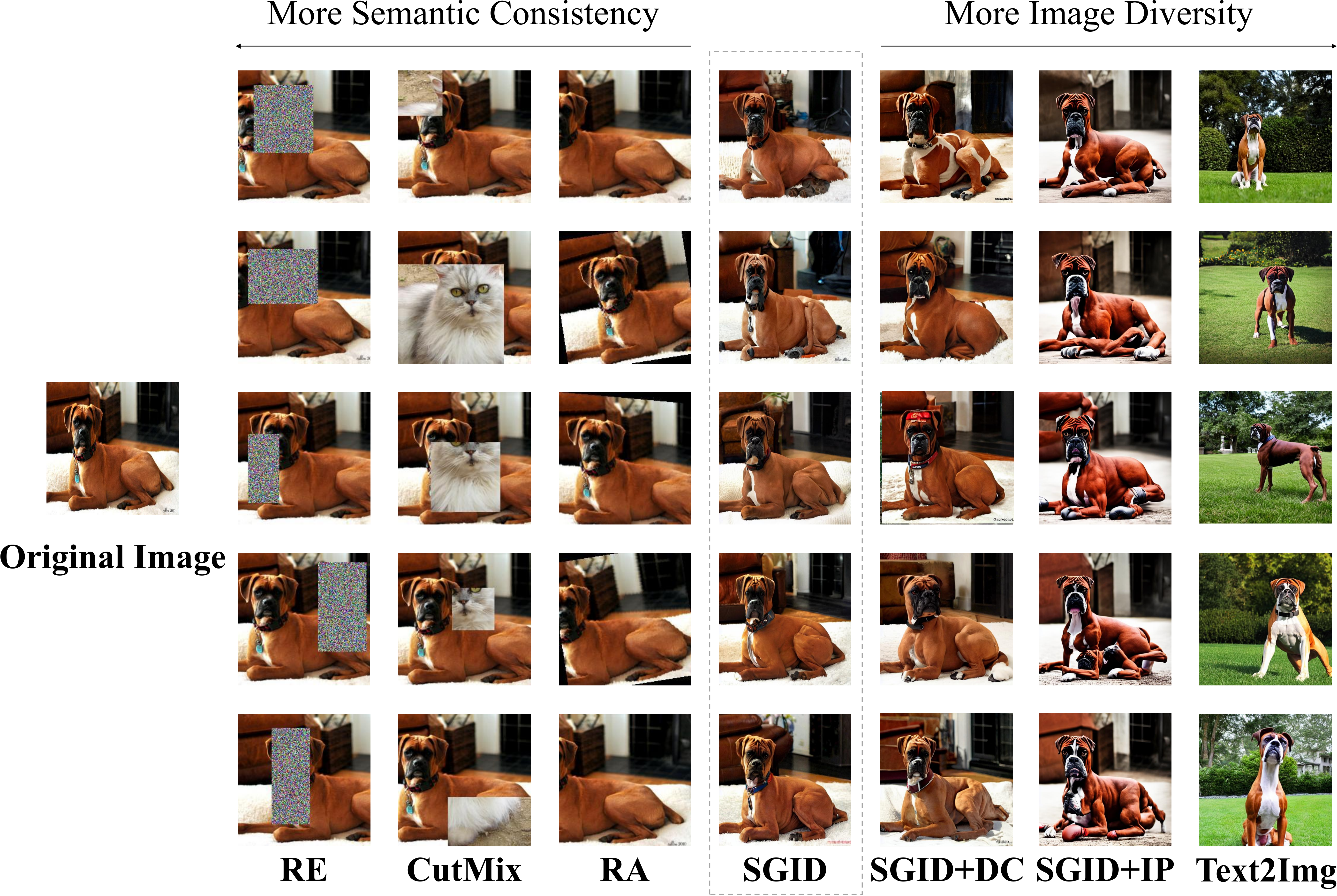}
  \caption{Augmented images by \proposed{} and other DA baselines on OxfordPets.}
  \label{fig:appendix-case-study-pets}
\end{figure*}

\begin{figure*}[!t]
  \centering 
  \includegraphics[width=0.8 \linewidth]{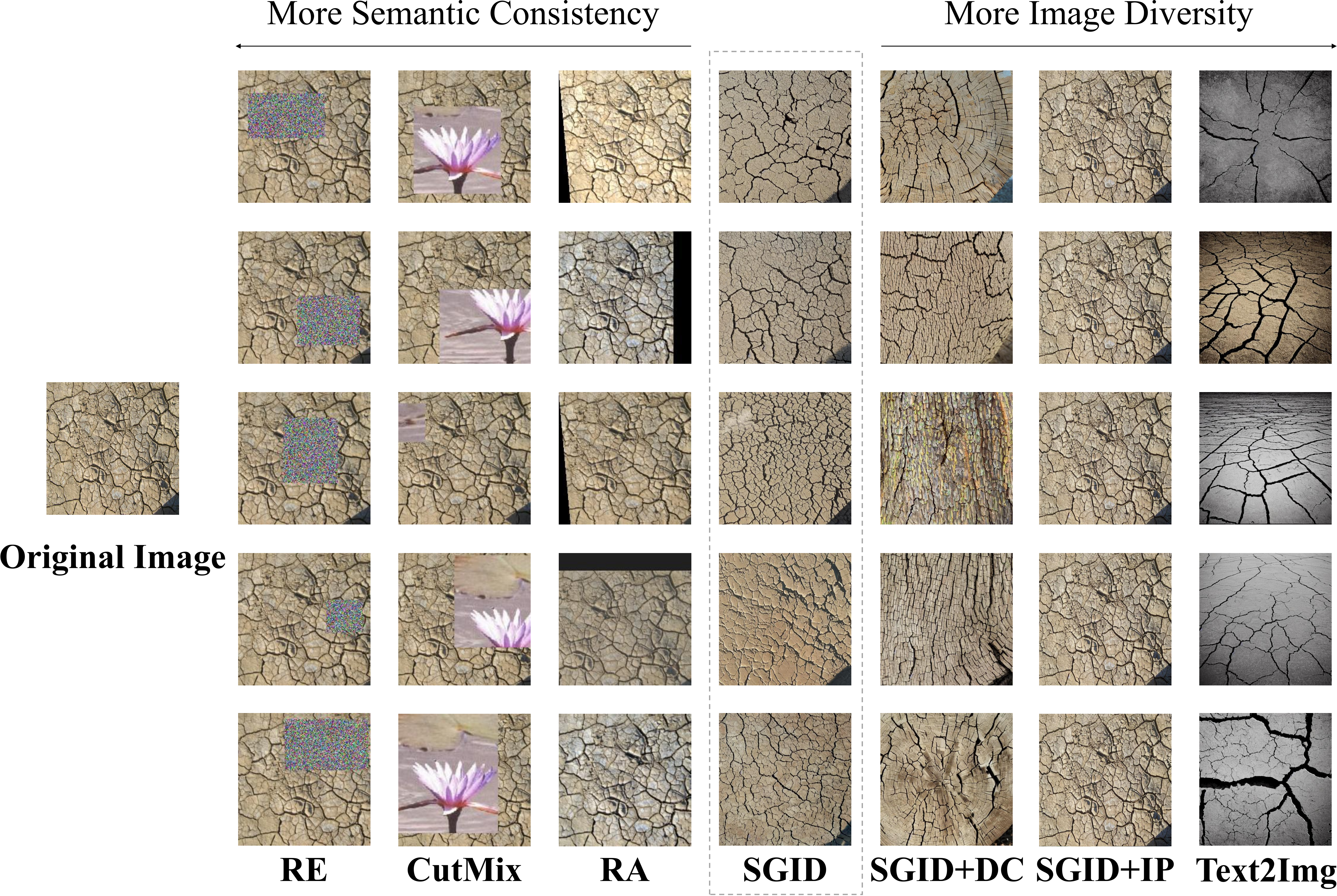}
  \caption{Augmented images by \proposed{} and other DA baselines on DTD.}
  \label{fig:appendix-case-study-DTD}
\end{figure*}

We show augmented images of \proposed{} and other image augmentation baselines on the following datasets, including CIFAR-100, CIFAR-10, Caltech101, Flowers102, OxfordPets, and DTD. 
The results are shown in 
Figure~\ref{fig:appendix-case-study-cifar100} - Figure~\ref{fig:appendix-case-study-DTD}.

\section{Other generative DA methods}\label{sec:appendix-generative-method}
\textit{GIF}~\cite{zhang2022expanding} optimizes the latent features of the original image in the semantic space and decodes the features into augmented images with new content~\cite{zhang2022expanding}. We compare \proposed{} with it in Appendix Sec.~\ref{tab:appendix-expand-results}. 
\textit{ALIA}~\cite{Dunlap2023DiversifyYV} generates captions for each original image, and use the captions across the datasets to language models and summarize them into some descriptions like image background. They use these descriptions as the captions or edit instructions to edit each original image. They conduct experiments on fine-grained datasets. We take the variants of this method as our baselines (\proposed{}+DiverseCaption and \proposed{}+InstructPix2Pix) because we have coarse-grained datasets and we summarize captions on each label, which is a more general setting.

\section{Datasets and Baselines}
\textbf{Datasets:} We evaluate the effectiveness of our proposed method on seven commonly used datasets, including three coarse-grained object classification datasets: {CIFAR-10}~\cite{krizhevsky2009learning}, {CIFAR-100}~\cite{krizhevsky2009learning}, {Caltech101} (Cal101) ~\cite{DBLP:conf/cvpr/LiFP04}, and four fine-grained object classification datasets: {Stanford Cars} (Cars)~\cite{krausecollecting}, {Flowers102} (Flowers)~\cite{nilsback2008automated}, {OxfordPets} (Pets)~\cite{parkhi2012cats} and texture classification~\cite{cimpoi2014describing} ({DTD}). 
Some other details are shown in Appendix Table~\ref{tab:dataset_setting}.

\section{Implementation Details}\label{sec:app-implementation}
\textbf{Image Generation}
We apply nucleus sampling and beam search to respectively generate $10$ captions by BLIP.
The caption length is between $5$ and $20$.
As for nucleus sampling, we use $p = 0.9$ by default as it is considered to be able to generate both fluent and diverse texts~\cite{li2022blip}.
As for beam search, we use $num\_beams$ = 3 by default.
The default ``CLIP-ViT-B/32'' model is used for calculating image-text similarity.
We apply the pre-trained ``stable-diffusion-v1-5'' model and generate one augmented image for each original image.~\footnote{SD v1-5 is mainly pre-trained on LAION-2B~\cite{Schuhmann2022LAION5BAO}.}
The $denoising\_steps$ is set as 100 for higher quality.
Empirically, we take $f(s^*) =-4 \cdot ({s^*}) ^{2}+ 2\cdot s^* + 1$ as the guidance mapping function.
We select the noise rate $n \in [0,1]$.\footnote{
The larger $n$ brings more variation.
We choose $n$ from $\{0.3, 0.5, 0.7\}$ to 
preserve image semantics and bring explicit changes in the background, position, etc.}
As for prompt weighting, We assign a weight of $1.50$ to the labels because it carries more accurate information for the original image, and a weight of $0.90$ to the caption to reduce the interference caused by the potential low-quality captions. 
We provide more details on training image classification models in Appendix Sec.~\ref{sec:app-implementation}.

\textbf{Image Classification:}
We train ResNet-50 (from scratch) from scratch for 300 epochs. The learning rate starts from 0.1 and is divided by 10 after the 150th and 225th epochs. The batch size is set to 256.
We use the SGD optimizer, and the momentum, weight decay are set to 0.9 and 5e-4. As for the pre-trained ViT (ImageNet-21k) and CLIP-ViT (LAION-2B), we use the framework and settings provided by HuggingFace~\cite{wolf-etal-2020-transformers} and OpenCLIP~\cite{Cherti2022ReproducibleSL}.
On the CIFAR-10 and CIFAR-100, the learning rate and training epoch are set to 5e-5 and 3. On the other datasets, we follow ~\cite{dosovitskiy2020image} and choose the learning rate based on the best performances among\{1e-4, 3e-4, 1e-3, 3e-3, 5e-5\} and the training epoch is set to 15.
All models are trained with the basic image transform methods: randomly performs horizontal flips and, for ResNet-50 (from scratch), takes a random resized crop with 32 $\times$ 32 on CIFAR-10 and CIFAR-100 datasets from images padded by 4 pixels on each side, 224 $\times$ 224 on the other datasets. 
Models are developed with PyTorch~\cite{NEURIPS2019_9015}. All experiments are conducted with NVIDIA A100 40GB and NVIDIA A100 80GB. More implementation details can be found in Appendix Sec.~\ref{sec:app-implementation}.

We adopt three visual backbones for image classification, including ResNet-50 (from scratch)~\cite{he2016deep},
WideResNet-28-10 (scratch)~\cite{zagoruyko2016wide}, 
ViT (ImageNet-21k) (\textit{vit-base-patch16-224-in21k};~\cite{dosovitskiy2020image}), and CLIP-ViT (LAION-2B)~\cite{Schuhmann2022LAION5BAO}. CLIP-ViT (LAION-2B) is pre-trained on LAION-2B~\cite{Schuhmann2022LAION5BAO}, its base model is ViT-B/32, and the amount of samples seen is 34B. Since ``stable-diffusion-v1-5'' we apply in this paper is pre-trained on LAION-2B and subset of ``laion-aesthetics v2 5+'' (600M), the pre-trained source of Stable Diffusion and CLIP-ViT (LAION-2B) are highly similar, thus CLIP-ViT (LAION-2B) is a strong backbone that demonstrates the performance of \proposed{}.
Among different DA methods, in order to ensure the fairness of the experiments, a grid search was performed with a fixed seed in the same hyperparameter search space, and the best result of each method is reported.
We develop and experiment based on the Pytorch~\cite{NEURIPS2019_9015} framework. 
All experiments are conducted with NVIDIA A100.

As for ResNet-50 (from scratch) and WideResNet-28-10 (scratch), we apply basic image transforms, including random horizontal flips and a random resized crop with 32 $\times$ 32 on CIFAR-10 and CIFAR-100 datasets (images are padded by 4 pixels on each side), and 224 $\times$ 224 on other datasets. 
We train these models from scratch for 300 epochs. 
Following~\cite{zhong2020random}, the learning rate starts from 0.1 and is divided by 10 after the 150$^\text{th}$ and 225$^\text{th}$ epoch.
The batch size is set to 256 for ResNet-50 (from scratch), and 64 for WideResNet-28-10 (scratch).
We use the SGD optimizer, and the momentum, weight decay are set to 0.9 and 5e-4 for both backbones.

As for ViT (ImageNet-21k), we use the pre-trained \url{https://huggingface.co/google/vit-base-patch16-224-in21k}{\textit{vit-base-patch16-224-in21k}} model and the image classification \url{https://github.com/huggingface/transformers/tree/main/examples/pytorch/image-classification}{framework} provided by HuggingFace~\cite{wolf-etal-2020-transformers}. 
Basic image transforms are applied, including random horizontal flips and a random resized crop with 224 $\times$ 224 to match the input dimensions of the pre-trained \textit{vit-base-patch16-224-in21k} model on all datasets. During model training, the batch size is set to 64 and the AdamW~\cite{loshchilov2018decoupled} optimizer is used. 
On CIFAR-10 and CIFAR-100, we use the default parameters from the framework, \ie{} the learning rate and training epoch are set to 5e-5 and 3. On the other datasets, we follow~\cite{dosovitskiy2020image} and choose the learning rate based on the best performances among\{1e-4, 3e-4, 1e-3, 3e-3\} and the training epoch is set to 15.

The hyper-parameters of baseline methods are as follows: 
\begin{itemize}
    \item For Random Erasing, we follow the \url{https://github.com/zhunzhong07/Random-Erasing}{official implementation} of ~\cite{zhong2020random} ($p=0.5$, $s_{h}=0.4$ and $r_{1}=0.3$);
    \item For CutMix, we follow the \url{https://github.com/clovaai/CutMix-PyTorch}{official implementation} of ~\cite{yun2019cutmix} ($\alpha=1$, cutmix probability is 0.5);
    \item For RandAugment, we use \url{https://github.com/pytorch/vision/blob/main/torchvision/transforms/autoaugment.py}{\textit{torchvision.transforms.RandAugment}} provided by PyTorch without changing any default parameters ($N=2$, $M=9$);
    \item For MoEx, we follow the \url{https://github.com/Boyiliee/MoEx}{official implementation} of \cite{li2021feature} ($\lambda=0.5$, $p=0.25$, PONO normalization).
\end{itemize}

\end{document}